\documentclass[preprint, 5p]{elsarticle}

\usepackage{hyperref}
\usepackage{amsmath,amssymb,amsfonts}
\usepackage{algorithmic}
\usepackage{textcomp}
\usepackage{xcolor}
\usepackage{tikz}
\usepackage{todonotes}
\usepackage{diagbox}
\usepackage{amsthm}

\usepackage[caption=false,font=footnotesize]{subfig}
\usetikzlibrary{shapes,decorations,arrows,calc,arrows.meta,fit,positioning}
\tikzset{
	-Latex,auto,node distance =0.50 cm and 0.50 cm,semithick, 
	state/.style ={circle, draw, minimum width = 0.8 cm, line width=1pt, inner sep =0pt },
	point/.style = {circle, draw, inner sep=0.04cm,fill,node contents={}},
	bidirected/.style={Latex-Latex,dashed},
	el/.style = {inner sep=2pt, align=left, sloped},
	association/.style={-,dashed}
}
\def\BibTeX{{\rm B\kern-.05em{\sc i\kern-.025em b}\kern-.08em
		T\kern-.1667em\lower.7ex\hbox{E}\kern-.125emX}}

\newtheorem{proposition}{Proposition}
\newproof{pf}{Proof}

\begin{document}

\begin{frontmatter}
	
	\title{Explaining the Model and Feature Dependencies \\ by Decomposition of the Shapley Value\tnoteref{t1}}
	\tnotetext[t1]{This project has received funding from the Flemish Government (AI
		Research Program) and from the FWO (`Artificial Intelligence (AI) for
		data-driven personalised medicine', G0C9623N and `Deep, personalized epileptic seizure detection’, G0D8321N) and Leuven.AI Institute.}

	\author[1]{Joran Michiels\corref{cor1}}
	\ead{joran.michiels@esat.kuleuven.be}
	\author[1,2]{Maarten De Vos}
	\ead{maarten.devos@kuleuven.be}
	\author[1]{Johan Suykens}
	\ead{johan.suykens@esat.kuleuven.be}
	
	\cortext[cor1]{Corresponding author}
	
	\affiliation[1]{organization={Stadius group in the Department of Electrical Engineering, KU Leuven University},
		addressline={Kasteelpark Arenberg 10},
		postcode={3001},
		city={Leuven},
		country={Belgium}}
	\affiliation[2]{organization={Department of Development and Regeneration,
			KU Leuven University},
		addressline={Herestraat 49},
		postcode={3000},
		city={Leuven},
		country={Belgium}}
	
	\begin{abstract}
		Shapley values have become one of the go-to methods to explain complex models to end-users. They provide a model agnostic post-hoc explanation with foundations in game theory: what is the worth of a player (in machine learning, a feature value) in the objective function (the output of the complex machine learning model). One downside is that they always require outputs of the model when some features are missing. These are usually computed by taking the expectation over the missing features. This however introduces a non-trivial choice: do we condition on the unknown features or not? In this paper we examine this question and claim that they represent two different explanations which are valid for different end-users: one that explains the model and one that explains the model combined with the feature dependencies in the data. We propose a new algorithmic approach to combine both explanations, removing the burden of choice and enhancing the explanatory power of Shapley values, and show that it achieves intuitive results on simple problems. We apply our method to two real-world datasets and discuss the explanations. Finally, we demonstrate how our method is either equivalent or superior to state-to-of-art Shapley value implementations while simultaneously allowing for increased insight into the model-data structure.
	\end{abstract}
%%Graphical abstract
%\begin{graphicalabstract}
	%\includegraphics{grabs}
%\end{graphicalabstract}
%%Research highlights
%\begin{highlights}
	%\item Research highlight 1
	%\item Research highlight 2
%\end{highlights}

\end{frontmatter}
	\section{Introduction}
	Black box machine learning methods have shown remarkable results on complex prediction tasks in the past years. Widespread adoption in practical settings is limited however. This is believed to be partly due to the lack of insights in the model's inner workings, making validating the model very difficult.  A popular choice to gain insight in the model's decisions are model agnostic local post-hoc explanations. These explain one sample (\textit{local}) and its corresponding prediction by sampling any (\textit{model agnostic}) already trained model (\textit{post-hoc}).
	
	One local post-hoc explanation, the \textbf{SH}apley \textbf{A}dditive ex\textbf{P}lanation (SHAP) \cite{lundberg2017unified}, has become increasingly popular in the last few years. Apart from its excellent Python module, this is likely due to the solid game-theoretic background of the method: other than the degree of approximation, nothing is left to the choice of the user, or so it seems. As has been recently noticed in related literature \cite{janzing2020feature,kumar2020problems,sundararajan2020many} the game-theoretic concept does not translate literally to machine learning models, leading to interpretation problems when features are dependent. It should be noted that similar problems can arise in other post-hoc explanations such as LIME \cite{ribeiro2016should}. They are however not covered in related papers (nor in our work). This is likely because uniqueness is one of the main selling points of SHAP, unlike other explanation methods.
	
	In our opinion, central to the understanding of these interpretation problems is the distinction between the model and the result (the model combined with the data), and the corresponding explanations. We qualitatively argue that different settings can have different requirements for local post-hoc explanations with a simple example (inspired by \cite{janzing2020feature} and \cite{sundararajan2020many}). Assume the following risk model $f_r$ and dependent features:
	\begin{equation}
		\begin{gathered}
			f_r([X_1,X_2]) = X_1 \\
			X_1, X_2\quad \text{binary} \\
			X_1 \not\!\perp\!\!\!\perp X_2
		\end{gathered}
	\end{equation}
	Let us imagine two different fictive settings with different feature definitions in Table \ref{intro_table}.	
	\begin{table}[h]
		\caption{Different settings with different features} \label{intro_table}
		\begin{center}
			\begin{tabular}{ c || c | c | c}
				& $X_1$ & $X_2$ & $f_r(\cdot)$ \\
				\hline
				\textbf{court} & recidivist & race & future recidivism \\
				\hline
				\textbf{hospital} & high HR & obesity & heart attack
				
			\end{tabular}
		\end{center}
	\end{table}
	Although the assumed data distribution is simplified, it is known that for both settings the features tend to be associated \cite{lockwood2015racial,rossi2015impact}. Given a certain sample, with SHAP we can explain its output by determining what effect every feature value has towards the output. In the first setting $f_r$ is an indicator for recidivism of a convict. By Article 6 of the European Court of Human Rights, a convict $\boldsymbol{x}$ can demand an explanation of the algorithm involved in the courts decision. In this case the explanation should attribute $f_r([\boldsymbol{x}_1,\boldsymbol{x}_2])$ completely to $\boldsymbol{x}_1$, showing the absence of racial bias in the \textit{model}. The second setting needs a different explanation. For patient $\boldsymbol{x}$ with a high heart attack risk, a good explanation should arguably attribute part of $f_r([\boldsymbol{x}_1,\boldsymbol{x}_2])$ to $\boldsymbol{x}_2$, since changing $\boldsymbol{x}_2$ by losing weight could decrease the heart rate, which in term reduces the odds of having a heart attack according to the model. Note that this last attribution is solely due to the dependency in the data. 
	
	Thus, although the model and data distribution are the same, different settings seem to require different kinds of explanations. The first explanation explains the model while the second tries to explain the model combined with the data, which we call the actual \textit{result}. Preferably, a good method should account for both settings. In this paper, we show that a SHAP value for a feature can be split up into a direct (model corresponding) attribution and an attribution via other dependent features, allowing the end-user (e.g. convict or patient) to easily grasp the explanation of \textit{both} the model and the result. Specifically, the contributions of this paper are:
	
	\begin{itemize}
		\item identification of the aforementioned distinction between explaining the model and explaining the result,
		\item an interpretation of a previously noticed decomposition \cite{heskes2020causal} as a unification of the two most common Shapley value implementations \cite{janzing2020feature, aas2019explaining},
		\item a new model-agnostic implementation of Shapley values with accompanying plots that provide enhanced insights in model and data,
		\item experiments demonstrating the correctness of our approach and its superiority over the current state-of-art \cite{kumar2021shapley} to quantify the effect of dependent features.
	\end{itemize}

	\section{Background}
	This section covers the necessary theoretical background of explanations based on Shapley values and their inherent problem with dependent features.
	\subsection{Shapley additive explanations} \label{axioms}
	SHAP is a post-hoc (i.e. after training) method to interpret the output of a complex model for a particular sample $\boldsymbol{x}$ (note the bold font) \cite{lundberg2017unified}. It constructs a local approximation $g$ of the original prediction model $f$ around $\boldsymbol{x}$ and is of the form:
	\begin{equation}
		g\left(X^{\prime}\right)=\phi_{0}+\sum_{i=1}^{M} \phi_{i} X_{i}^{\prime} \quad \text{with} \quad X^{\prime} \in\{0,1\}^{M} \ \ \phi_i \in \mathbb{R},
	\end{equation}
	with $X^\prime$ the so-called simplified input of the original input $X$, and $\phi_{i}$ the actual SHAP values or feature attributions. To transform $X'$ to $X$ a local mapping function $h_{\boldsymbol{x}}$ is defined. In general any mapping $X=h_{\boldsymbol{x}}\left(X^{\prime}\right)$ can be utilized, in practice $X_i^\prime$ will be defined as an indicator whether the particular input feature is present or not, i.e. for particular input $\boldsymbol{x}$, $X_{i}^{\prime} = \textbf{1}\left[X_{i} = \boldsymbol{x}_{i}\right]$. In this case $M$ is equal to the number of features and the explanation effectively becomes an additive function on feature presence.
	
	To select reasonable feature attributions $\phi_{i}$ for the explanation model, three properties are imposed.
	\begin{enumerate}
		\item \textbf{Local accuracy}:
		\begin{equation}
			f\left(\boldsymbol{x}\right)= g\left(\boldsymbol{x}^{\prime}\right)
		\end{equation}
		i.e. at the sample to explain $\boldsymbol{x}$, the explanation matches the original model.
		\item \textbf{Missingness}:
		\begin{equation}
			X_{i}^{\prime}=0 \implies \phi_{i}=0,
		\end{equation}
		i.e. if the original input $X$ already had a missing feature, its attribution is always zero. The equivalent form,
		\begin{equation}\label{eq:missingness}
			f(h_{\boldsymbol{x}}(x^{\prime} \cup i)) = f(h_{\boldsymbol{x}}(x^{\prime} \backslash i))  \ \forall x^\prime \implies \phi_{i}=0
		\end{equation}
		with $x^{\prime} \cup i$ meaning $x_{i}^{\prime}=1$ and with $x^{\prime} \backslash i$ meaning $x_{i}^{\prime}=0$, is clearer.
		
		\item \textbf{Consistency}:
		
		For any two models $f_1$ and $f_2$:
		\begin{equation}
			\begin{gathered}
				f_{1}\left(h_{\boldsymbol{x}}\left(x^{\prime}\right)\right)-f_{1}\left(h_{\boldsymbol{x}}\left(x^{\prime}\backslash i\right)\right)  \\ \geq f_{2}\left(h_{\boldsymbol{x}}\left(x^{\prime}\right)\right)-f_{2}\left(h_{\boldsymbol{x}}\left(x^{\prime} \backslash i\right)\right)\ \forall x^{\prime} \in\{0,1\}^{M} \\
				\implies
				\phi_{i}\left(f_{1},\boldsymbol{x}\right) \geq \phi_{i}(f_{2},\boldsymbol{x})
			\end{gathered}
		\end{equation}
		In words, this says that ``if a model changes so that some simplified
		input's contribution increases or stays the same regardless of the other inputs, that input's attribution
		should not decrease'' \cite{lundberg2017unified}.
	\end{enumerate}
	Only one\footnote{\cite{sundararajan2020many} claim other properties are necessary for uniqueness.} possible set of feature attributions satisfies these properties \cite{lundberg2017unified}:
	\begin{equation} \label{eq:shap}
		\phi_{i}(f, \boldsymbol{x})=\frac{1}{M!}\sum_{R} f\left(h_{\boldsymbol{x}}\left(S^R \cup \{i\}\right)\right)-f\left(h_{\boldsymbol{x}}\left(S^R\right)\right),
	\end{equation}
	where $R$ is a permutation of the ordering of simplified inputs and $S^R$ the set of simplified inputs preceding $i$ in this ordering. To make the formula less cluttered, we use $\left(S^R\right)$ to denote $\left(\left[X_{S^R}^\prime = 1, X_{\bar{S}^R}^\prime =0\right]\right)$. The values $\phi_{i}$ are known as Shapley values in cooperative game theory \cite{shapley1953value}. In practice, when the simplified inputs correspond to feature presence, $\phi_{i}$ is the average over all orderings of the change in model output when adding feature $i$ to the preceding, already `known' features. Equation \eqref{eq:shap} requires the output of the model when features are missing: $f\left(h_{\boldsymbol{x}}\left(X^{\prime}\right)\right)$ when $X^\prime \neq \boldsymbol{x}^\prime$. This involves some empirical expectation of $f\left(h_{\boldsymbol{x}}\left(X^{\prime}\right)\right)$ over a set of training samples. What expectation to choose - conditioning on the missing features or not - will appear to be a controversial topic.
	
	\subsection{Missing features}
	The original SHAP paper \cite{lundberg2017unified} assumed a seemingly natural way to express the expectation of the model output with missing features: conditioning on known features $S$,
	\begin{equation}\label{eq:ces}
		f\left(h_{\boldsymbol{x}}\left(X^{\prime}\right)\right) =E\left[f(X) | X_S = \boldsymbol{x}_{S}\right]. \ \ \text{(conditional SHAP)}
	\end{equation} In practice, the exact data distribution is unknown. Therefore \cite{lundberg2017unified} proposed to use independent features. Then the expectation can be easily approximated by averaging over $K$ background samples $x^k$:
	\begin{align}
		\label{eq:ies}    E\left[f(X) | X_S = \boldsymbol{x}_{S}\right] & \approx E\left[f\left(\left[\boldsymbol{x}_S, X_{\bar{S}}\right] \right)\right]                                                                                    \\
		& \approx \frac{1}{K} \sum_{k}f\left(\left[\boldsymbol{x}_S, x^k_{\bar{S}}\right] \right),
	\end{align} with missing features $\bar{S}$.
	A later contribution by the same authors \cite{lundberg2018consistent} focussed on avoiding the feature independence assumption by exploiting the structure of tree models. Part of the community also followed this direction: \cite{aas2019explaining, aas2021explaining} estimate the conditional data distribution $p\left(X_{\bar{S}} | X_{S}\right)$ to better approximate \eqref{eq:ces} and \cite{takeishi2019shapley} uses the data distribution assumed by probabilistic PCA to `exactly' compute the SHAP value.

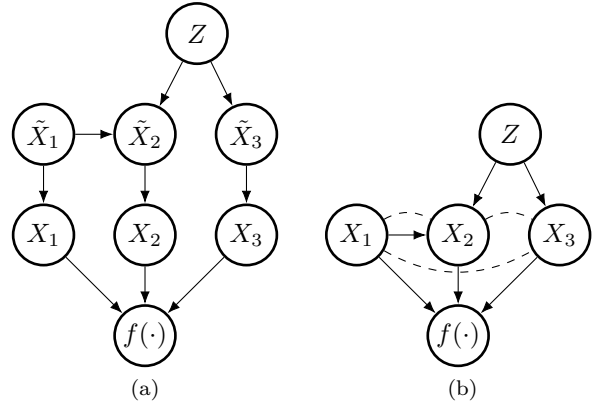
\begin{figure}
		\centering
		\subfloat[]{
			\begin{tikzpicture}
				% x node set with absolute coordinates
				\node[state] (x2) at (0,0) {$X_2$};
				
				% y node set relative to x.
				% Locations can be:
				% right,left,above,below,
				% above left,below right, etc
				\node[state] (x1) [left =of x2] {$X_1$};
				\node[state] (x3) [right =of x2] {$X_3$};
				\node[state] (x1t) [above =of x1] {$\tilde{X}_1$};
				\node[state] (x2t) [above =of x2] {$\tilde{X}_2$};
				\node[state] (x3t) [above =of x3] {$\tilde{X}_3$};
				\node[state] (z)  [xshift=-0.65cm, above =of x3t] {$Z$};
				
				\node[state] (f) [below =of x2] {$f(\cdot)$};
				
				% Directed edge
				\path (x1t) edge (x1);
				\path (x2t) edge (x2);
				\path (x3t) edge (x3);
				
				\path (x1) edge (f);
				\path (x2) edge (f);
				\path (x3) edge (f);
				
				\path (x1t) edge (x2t);
				
				\path (z) edge (x2t);
				\path (z) edge (x3t);

			\end{tikzpicture}
			\label{causal_int}}
		\hfil
		\subfloat[]{\begin{tikzpicture}
				% x node set with absolute coordinates
				\node[state] (x2) at (0,0) {$X_2$};
				
				% y node set relative to x.
				% Locations can be:
				% right,left,above,below,
				% above left,below right, etc
				\node[state] (x1) [left =of x2] {$X_1$};
				\node[state] (x3) [right =of x2] {$X_3$};
				\node[state] (f) [below =of x2] {$f(\cdot)$};
				\node[state] (z)  [xshift=-0.65cm, above =of x3] {$Z$};
				
				% Directed edge
				
				\path (x1) edge (f);
				\path (x2) edge (f);
				\path (x3) edge (f);
				
				\path (x1) edge (x2);
				
				\path (z) edge (x2);
				\path (z) edge (x3);
				
				\path[association] (x1) edge[bend left=30] (x2);
				\path[association] (x2) edge[bend left=30] (x3);
				\path[association] (x1) edge[bend right=30] (x3);
				
			\end{tikzpicture}
			\label{causal_all}}
		\caption{Example causal structures assumed by respectively interventional SHAP \cite{janzing2020feature} (a) and causal Shapley values \cite{heskes2020causal} (b). In (a) the $\tilde{X}_i$ are the real world features and $X_i$ the model inputs. This distinction is not made in (b). For both cases $Z$ is a latent confounder. The dotted lines are the observed statistical dependencies which are taken into account by conditional SHAP, and the new method proposed in this paper.}
		\label{causal}
	\end{figure}

	Some recent papers have tried to advocate a different estimation of the model output when features are missing. \cite{janzing2020feature, heskes2020causal, frye2020asymmetric} analyse the missing feature problem from a causal perspective. The conditioning on known features is argued to be \textit{interventional}, instead of \textit{observational} (as was implicitly assumed in conditional SHAP). Computing interventional conditional distributions requires a causal graph over the features. \cite{janzing2020feature} distinguishes between the true real world features $\tilde{X}$ and the model inputs $X$. There can be causal relations between the former but not between the latter. \cite{heskes2020causal} does not make this distinction and the resulting SHAP values are arguably the true causal output effects. The authors therefore deemed their explanations \textit{causal Shapley values.} To illustrate the difference with the approach of \cite{janzing2020feature}, a causal graph for a model with three input features is shown in Figure \ref{causal} for both approaches.
	
	Contrary to causal Shapley values \cite{heskes2020causal} the method of \cite{janzing2020feature} does not require any knowledge of the causal relations between the (real world) features. The authors show that the particular causal structure (e.g Figure \ref{causal_int}) leads to a model output equivalent to the left hand side of \eqref{eq:ies}: 
	\begin{equation}\label{eq:rbs}
		f\left(h_{\boldsymbol{x}}\left(X^{\prime}\right)\right) =E\left[f\left(\left[\boldsymbol{x}_S, X_{\bar{S}}\right]\right)\right].  \ \text{(interventional SHAP)}
	\end{equation}
	Without the weight of the coalition, the contribution for one permutation $R$, of feature $i$ then becomes
	\begin{equation}\label{eq:intSHAPcont} \phi_{i, S^R} = E\left[f\left(\left[\boldsymbol{x}_{S^R},\boldsymbol{x}_i, X_{\bar{S}^R_i}\right]\right)\right] - E\left[f\left(\left[\boldsymbol{x}_{S^R}, X_{\bar{S}^R}\right]\right)\right] 
	\end{equation}
	instead of 
	\begin{equation}
		\begin{gathered}
			\phi_{i, S^R} = E\left[f\left(\left[\boldsymbol{x}_{S^R},\boldsymbol{x}_i, X_{\bar{S}^R_i}\right]\right) | \boldsymbol{x}_{S^R}, \boldsymbol{x}_i\right] \\ - E\left[f(\left[\boldsymbol{x}_{S^R}, X_{\bar{S}^R}\right]) | \boldsymbol{x}_{S^R}\right]
		\end{gathered} 
	\end{equation}
	for conditional SHAP, with $S^R_i$ the set $S^R$ including feature $i$. In short, interventional SHAP breaks the dependencies between $X_i$ and $X_{\bar{S}^R_i}$, while conditional SHAP allows for them to have an effect. It is important to note that interventional SHAP keeps the dependencies within $X_{\bar{S}^R_i}$ \cite{janzing2020feature}.
	
	\cite{sundararajan2020many} uses the axiomatic approach to study the attribution problem. They test interventional and conditional SHAP against a list of desirable properties. Contrary to the properties imposed in Section  \ref{axioms}, these are at the level of the model $f$ not at the level of the model approximations ($f\left(h_{\boldsymbol{x}}\left(X^{\prime}\right)\right)$ when $X^\prime \neq \boldsymbol{x}^\prime$), e.g. missingness \eqref{eq:missingness} becomes `dummy'.
	
	\textbf{Dummy}:
	\begin{equation}\label{eq:missing_mod}
		f([x_i, x_{\backslash i}]) = f([d, x_{\backslash i}]) \ \forall x, d \implies \phi_{i}=0
	\end{equation}
	The authors show that conditional SHAP fails this property (and others), while interventional SHAP does not.
	
	\section{Related work} 
	
	Recently \cite{chen2020true} also suggested the same distinction as described in our introduction and motivate each of the approaches with a use case. Interventional SHAP seems the method of choice when a loan applicant wants to see what features directly increase their likelihood of being granted a loan. If we want to know more about the natural mechanism of gene expression in a patient with a type of blood cancer, the conditional SHAP value is preferred and identified more of the true underlying variables. They call this approach `true to the data'. In contrast with this paper, no attempt to unify both types of SHAP values was undertaken.
	
	Similar to our contribution, \cite{heskes2020causal} decomposes the causal Shapley value into a direct and indirect effect and applies the decomposition to a theoretical example to show how different causal graphs result in different Shapley values. They do not provide a possible implementation and neither do they motivate using the decomposition in practice. Here we interpret this decomposition as a unification of conditional and interventional SHAP, and provide an implementation and illustrations to increase the understanding of the method. Furthermore we compare it with other state-of-the art Shapley implementations.

	Shapley residuals \cite{kumar2021shapley} can capture limitations of regular Shapley value implementations. They are represented by residual vectors $r_i$ measuring how distant all contributions for specific $S$ ($\phi_{i, S}$) are from the contributions of the nearest inessential game. In an inessential game every player (in case of machine learning a feature value) contributes a fixed amount to the objective regardless of  $S$. As shown in \cite{kumar2021shapley} the contributions of the nearest inessential game are the actual Shapley values ($\phi_{i, S} = \phi_i$). In case of conditional SHAP value, Shapley residuals quantify the effect of dependent features. Specifically, it is argued that the norm of $r_i$ relates to the effect of dependencies between feature $i$ and other features on the SHAP value. For interventional SHAP values, the residuals capture effects of feature interactions in the model (e.g. of the term $X_1X_2$ in model $f(X_1, X_2) = X_1 + X_2 + X_1X_2$). Explaining such feature interaction (also see Shapley-Interaction interaction indices \cite{sundararajan2020shapley}) is not the interest of this paper.
	
	\cite{wang2021shapley} also design a Shapley value implementation that provides a more involved explanation when dependencies between input features exist. In fact, they assign a contribution to every causal link (or edge in causal graph) while still adhering to the fundamental axioms described in Section \ref{axioms}. A connection with conditional SHAP is made but interventional SHAP values are not discussed. Similarly to causal Shapley values \cite{heskes2020causal} a causal graph over the features is required.
	
	Another important difference between interventional and conditional SHAP values is that the former evaluates the machine learning model on out of distribution samples. Intervening on feature values results in samples which do not represent the underlying data distribution (also called off-manifold). It is shown that this allows for adversarial attacks that can manipulate explanations to hide unwanted model biases \cite{slack2020fooling}. While we acknowledge this issue, interventions still provide important insight into model as motivated in our introduction (e.g. for bias detection) and in \cite{chen2020true}. For more information on the off-manifold problem we direct the reader to \cite{frye2020shapley,yeh2022threading}. In this paper, the focus will be on the meaningful usage and desired theoretic properties of SHAP values, and not on their robustness or efficiency.

	\section{Explaining model and result} \
	The properties defined in \cite{sundararajan2020many} (such as \eqref{eq:missing_mod}) quantify the model correspondence of the explanations. Conditional SHAP, not adhering to many of these properties, sacrifices this model correspondence for more result-centric feature attributions. These latter attributions arguably show the actual statistical relations between inputs and output (or result). Note that the relations may not be fully present in the data. The model fitting enforces some domain and/or data knowledge. Also, the complexity of the model might be limited for computational reasons, meaning that the `true' relations in the data are not uncovered. On this point we do not completely agree with \cite{chen2020true}. They mention that conditional SHAP values are true to the data and interventional SHAP values true to the model. As we argued, conditional SHAP values are both true to the data \textit{and} true to the model.
	
	Causal Shapley values \cite{heskes2020causal} show the true \textit{causal} relations between inputs and output, and thus also explain the result. The big difference is that conditional SHAP relies on association (or statistical dependence) and the causal Shapley value on causation. For example in Figure \ref{causal_all} $X_2$ and $X_3$  have a common cause $Z$ and will be associated (note the dotted line) without a causal relationship. We agree with \cite{heskes2020causal} that causal relations are more intuitive than observed statistical dependencies. However, in practice causal knowledge about the input features is often unavailable, hence the use of conditional SHAP in the rest of this paper.
	
	It should be clear that in the introductory example interventional SHAP values are preferred in court; and in the hospital setting the conditional version would be better. Conditional SHAP also finds the feature's attribution through other dependent features. We can show that the feature's attribution can be split up into its `direct' interventional effect and its `indirect' dependent effect. This decomposition of the SHAP value can be used in both the court and hospital setting.

	\subsection{Decomposition of the SHAP value}\label{decomp}
	
	Since our interest is decomposing the output effect into an interventional effect of $\boldsymbol{x}_i$ and its effect via the unknown features $\bar{S}_i$, we determine the effect of intervening on $\boldsymbol{x_i}$ when the features in $S$ are already known. I.e. the effect of setting $X_i = \boldsymbol{x_i}$ (breaking $X_i$'s dependencies) on $f_S(\left[X_i, X_{\bar{S}_i}\right]) = f(\left[\boldsymbol{x}_S,X_i, X_{\bar{S}_i} \right])$. This is equal to:
	\begin{equation}
		\begin{gathered}
		E\left[f_S\left(\left[\boldsymbol{x}_i, X_{\bar{S}_i}\right]\right)\right]
		 - E\left[f_S(X_{\bar{S}})\right] \\
		 = E\left[f\left(\left[\boldsymbol{x}_{S},\boldsymbol{x}_i, X_{\bar{S}_i}\right]\right) | \boldsymbol{x}_{S}\right]
		  - E\left[f(\left[\boldsymbol{x}_{S}, X_{\bar{S}}\right]) | \boldsymbol{x}_{S}\right]
		\end{gathered} 
	\end{equation}
	This effect can be interpreted as a part of the contribution for one permutation $R$ in case of the conditional SHAP value:
	\begin{equation}\label{eq:split}
		\begin{split}
			\phi_{i, S^R}(f, \boldsymbol{x}) = & E\left[f(X) | X_{S^R} = \boldsymbol{x}_{S^R}, Z_i = \boldsymbol{x}_i\right] \\ 
			& - E\left[f(X) | X_{S^R} = \boldsymbol{x}_{S^R}\right] \\
			= & E\left[f\left(\left[\boldsymbol{x}_{S^R},\boldsymbol{x}_i, X_{\bar{S}^R_i}\right]\right) | \boldsymbol{x}_{S^R}, \boldsymbol{x}_i\right]\\
			& - E\left[f(\left[\boldsymbol{x}_{S^R}, X_{\bar{S}^R}\right]) | \boldsymbol{x}_{S^R}\right] \\
			= & \Bigl( E\left[f\left(\left[\boldsymbol{x}_{S^R},\boldsymbol{x}_i, X_{\bar{S}^R_i}\right]\right) | \boldsymbol{x}_{S^R}\right] \\
			& - E\left[f(\left[\boldsymbol{x}_{S^R}, X_{\bar{S}^R}\right]) | \boldsymbol{x}_{S^R}\right] \Bigr)   \\
			&+  \Bigl( E\left[f\left(\left[\boldsymbol{x}_{S^R},\boldsymbol{x}_i, X_{\bar{S}^R_i}\right]\right) | \boldsymbol{x}_{S^R}, \boldsymbol{x}_i\right]\\ 
			&- E\left[f\left(\left[\boldsymbol{x}_{S^R},\boldsymbol{x}_i, X_{\bar{S}^R_i}\right]\right) | \boldsymbol{x}_{S^R}\right] \Bigr) \\
			= & \phi_{i, S^R, int}(f, \boldsymbol{x}) + \phi_{i, S^R, dep}(f, \boldsymbol{x}).
		\end{split}
	\end{equation}
	
	We have arrived at an equivalent decomposition as in \cite{heskes2020causal} but without causal operations. From \eqref{eq:split} we see that the contribution of a feature $i$ (for a permutation $R$) is the sum of an interventional output effect of $\boldsymbol{x}_i$ conditioned on $\boldsymbol{x}_{S^R}$ and the output effect of knowing $\boldsymbol{x}_i$ on the unknown variables, again conditioned on $\boldsymbol{x}_{S^R}$. Notice the similarity between $\phi_{i, S^R, int}(f, \boldsymbol{x})$ and interventional SHAP \eqref{eq:intSHAPcont}: the dependencies between $X_i$ and $X_{\bar{S}^R_i}$ are still broken. Consequently, the causal motivation behind interventional SHAP still holds and this decomposition successfully connects the interventional and conditional approach to dealing with missing features.
	
	Note that above interpretation also holds for the  Shapley value since that is an average over all $\phi_{i, S^R}(f, \boldsymbol{x})$. Thus, the conditional SHAP value can be split up into two SHAP value parts. Although these parts are not real SHAP values (there respective sums do not equal the original output values, i.e. no local accuracy), they arguably provide both a model and a result explanation. 
	
	As explained in the related work, Shapley residuals \cite{kumar2021shapley} for a particular model $f$ and feature  $i$ can be interpreted as the geometric difference between the contributions $\phi_{i,S}$ for all $S$ and the SHAP value $\phi_{i}$, i.e. $r_i \in \mathbb{R}^{|S|}$ and $r_{i, S} = \phi_{i,S} - \phi_{i}$. Thus, they define a decomposition of similar form as Equation (\ref{eq:split}) which was not discussed in \cite{kumar2021shapley}, namely:
	\begin{equation}\label{eq:decomposition_residuals}
		\phi_{i, S^R}(f, \boldsymbol{x}) = \phi_i(f, \boldsymbol{x}) + r_{i, S^R}
	\end{equation}
	This allows us to claim the following proposition. 
	
	\begin{proposition}\label{prop_residual}
		The associated SHAP part of the Shapley residual is always zero.
	\end{proposition}
	\begin{pf}
	 The SHAP part can be found by averaging \eqref{eq:decomposition_residuals} over all $R$:
	\begin{equation}
	 	\frac{1}{M!}\sum_R \phi_{i, S^R}(f, \boldsymbol{x}) = \phi_i(f, \boldsymbol{x}) + \frac{1}{M!}\sum_R r_{i, S^R}.
	\end{equation}
	Since the left hand side is exactly equal to the SHAP value we arrive at:
	\begin{equation}
			 \frac{1}{M!}\sum_R r_{i, S^R} = 0.
	\end{equation}
    \qedsymbol
	\end{pf}
	Because the associated SHAP part will always be zero, the concept of Shapley residuals can not be used to split the SHAP value into a sum of parts.
	\subsection{Theoretical validation} \label{validation}
	
	It is easily proven that the interventional part passes the dummy property \eqref{eq:missing_mod}, underlining its model correspondence. Also, for an independent distribution the interventional SHAP part coincides with the interventional SHAP value.
	\begin{proposition} \label{prop_dummy}
		Interventional SHAP part passes the \\ dummy property \eqref{eq:missing_mod}, i.e.
		\begin{equation}
			f([x_i, x_{\backslash i}]) = f([d, x_{\backslash i}]) \ \forall x, d \implies \phi_{i, int}=0
		\end{equation}
	\end{proposition}
	\begin{pf}	
		Take feature $i$ to be a missing feature:
		\begin{equation}\label{eq:missing_feature}
			f([x_i, x_{\backslash i}]) = f([d, x_{\backslash i}])  \quad \forall x, d.
		\end{equation}
		Its feature value contribution for permutation $R$,
		\begin{equation}
			\begin{split}
				\phi_{i, S^R, int}(f, \boldsymbol{x}) = & E\left[f\left(\left[\boldsymbol{x}_{S^R},\boldsymbol{x}_i, X_{\bar{S}^R_i}\right]\right) | \boldsymbol{x}_{S^R}\right] \\
				& - E\left[f(\left[\boldsymbol{x}_{S^R}, X_{\bar{S}^R}\right]) | \boldsymbol{x}_{S^R}\right] \\
				= & E\left[f\left(\left[\boldsymbol{x}_{S^R},\boldsymbol{x}_i, X_{\bar{S}^R_i}\right]\right) | \boldsymbol{x}_{S^R}\right] \\ 
				& - E\left[f\left(\left[\boldsymbol{x}_{S^R}, X_i, X_{\bar{S}^R_i}\right]\right) | \boldsymbol{x}_{S^R}\right]
			\end{split}
		\end{equation}
		is zero since both terms are equal by \eqref{eq:missing_feature}. The actual SHAP part, an average over the former differences, is thus also zero. \qedsymbol
	\end{pf}
	
	\begin{proposition}\label{prop_additive}
		Our proposed decomposition \eqref{eq:split} divides the model output effect among the model's additive components. Writing $f(X) = \sum_{A \subseteq F} f(X_A)$ then
		\begin{equation}
			\begin{gathered}
			\phi_{i, int}(f, \boldsymbol{x}) = \phi_{i, int}\left(\sum_{A \subseteq F|i \in A} f_A, \boldsymbol{x}\right) \\
			\phi_{i, S^R, dep}(f, \boldsymbol{x}) = \phi_{i, S^R, dep}\left(\sum_{A \subseteq F|S_R \subseteq A} f_A, \boldsymbol{x}\right)
			\end{gathered}
		\end{equation}
	\end{proposition}
	\begin{pf}
	Using Equation \eqref{eq:split} and knowing that the expectation of a sum of variables is equal to the sum of its individual expectations, we find that parts of the additive function cancel each other resulting in the claimed properties.	
	\qedsymbol	
	\end{pf}
	We can gain extra intuition by considering a linear model: $f=\sum_i a_i X_i$. In this case, the interventional SHAP values $\psi_i$, interventional SHAP parts $\phi_{i,int}$ and dependent SHAP parts $\phi_{i, dep}$ for a sample $\boldsymbol{x}$ simplify to:
	\begin{equation}\label{eq:linear_int}
		\psi_i = a_i(\boldsymbol{x}_i - E[X_i]) = f(\boldsymbol{x}) - f\left(\left[\boldsymbol{x}_{\backslash i}, E[X_i]\right]\right),
	\end{equation}
	\begin{equation}\label{eq:linear_int_part}
		\begin{split}
			\phi_{i, int} = & \frac{1}{M!}\sum_R a_i(\boldsymbol{x}_i - E[X_i|\boldsymbol{x}_{S^R}]) \\
			= &\frac{1}{M!}\sum_R  f(\boldsymbol{x}) - f\left(\left[\boldsymbol{x}_{\backslash i}, E[X_i| \boldsymbol{x}_{S^R}]\right]\right),
		\end{split}
	\end{equation}
	\begin{equation}
		\phi_{i, dep} = \frac{1}{M!}\sum_R \sum_{j \in \bar{S}_i^R} a_j(E[X_j|\boldsymbol{x}_{S^R}, \boldsymbol{x}_i] - E[X_j| \boldsymbol{x}_{S^R}]).
	\end{equation}
	
	First of all, notice that the interventional SHAP value corresponds with the change in model output when setting the respective feature to its average value. The interventional SHAP part, on the other hand, is the average over all permutations of the change in model output when intervening with a conditional mean. The conditioning is on the feature values appearing before $i$ in the permutation. In one of the experiments below we will argue why the latter mean can be more realistic. Note that the above observations are only exact for a linear machine learning model. Lastly, as expected from Proposition \ref{prop_additive}, $\phi_{i, int}$ only depends on $X_i$'s coefficient, while the dependent SHAP part $\phi_{i, dep}$ depends on all the other coefficients.

	\subsection{Implementation}
	
	Computing the SHAP values from \eqref{eq:shap} exactly is only feasible if the model has a few inputs, since it requires a sum over all possible orderings of inputs. To efficiently compute the SHAP value for models with more inputs, \cite{lundberg2017unified} proposes Kernel SHAP which interprets the computation of Equation \eqref{eq:shap} as performing a weighted linear regression. This is shown to be more sample-efficient than straightforwardly sampling the sum in \eqref{eq:shap}. Computation of the SHAP parts can not be expressed as a weighted linear regression.  Hence, to calculate the interventional SHAP part, we use a more rudimentary sampling approach as proposed by \cite{vstrumbelj2014explaining}. 
	
	The complete procedure for a sample $\boldsymbol{x}$ is as follows:
	\begin{enumerate}
		\item obtain or estimate the conditional data distribution $p\left(X_{\bar{S}} | X_{S}\right)$ ;
		\item compute the conditional SHAP values by applying Kernel SHAP, estimating \newline $E\left[f(X) | X_S = \boldsymbol{x}_{S}\right] \approx \frac{1}{K_1} \sum_{k}f\left(\left[\boldsymbol{x}_S, x^k_{\bar{S}}\right]\right)$ with $x^k_{\bar{S}}$ sampled from $p\left(X_{\bar{S}} | X_{S} = \boldsymbol{x}_S\right)$;
		\item compute the interventional SHAP parts:
		\begin{enumerate}
			\item sample $K_2$ feature permutations;
			\item for each permutation $R$ sample single data point $x^1_{\bar{S}^R}$ from $p\left(X_{\bar{S}^R} | X_{S^R} = \boldsymbol{x}_{S^R}\right)$;
			\item $\phi_{i, int} \approx \newline \frac{1}{K_2} \sum_{R}f\left(\left[\boldsymbol{x}_{S^R}, \boldsymbol{x}_i, x^1_{\bar{S}^R_i}\right]\right) - f\left(\left[\boldsymbol{x}_{S^R}, x^1_{\bar{S}^R}\right]\right)$; 
		\end{enumerate}
		\item compute the dependent SHAP parts by subtracting the interventional parts from the SHAP values.
	\end{enumerate}
	
	Practically, $K_1$ and $K_2$ are increased until sufficient convergence is achieved. Due to the decreased sample efficiency \cite{lundberg2017unified}, we advise $K_2$ to be taken bigger than $K_1$.
	
	The above procedure can be applied for any conditional data distribution which can be sampled. In our experiments we used a multivariate Gaussian distribution and a Gaussian copula (as proposed by \cite{aas2019explaining}). The complete algorithm in the case of a multivariate Gaussian data distribution is available in the Appendix. Like the original Kernel SHAP, our implementation is model agnostic and thus can be used with any model $f$ including neural networks and support vector machines.
	
	\section{Experiments}
	
\begin{figure}
		\centering
		\includegraphics[width=0.95\linewidth]{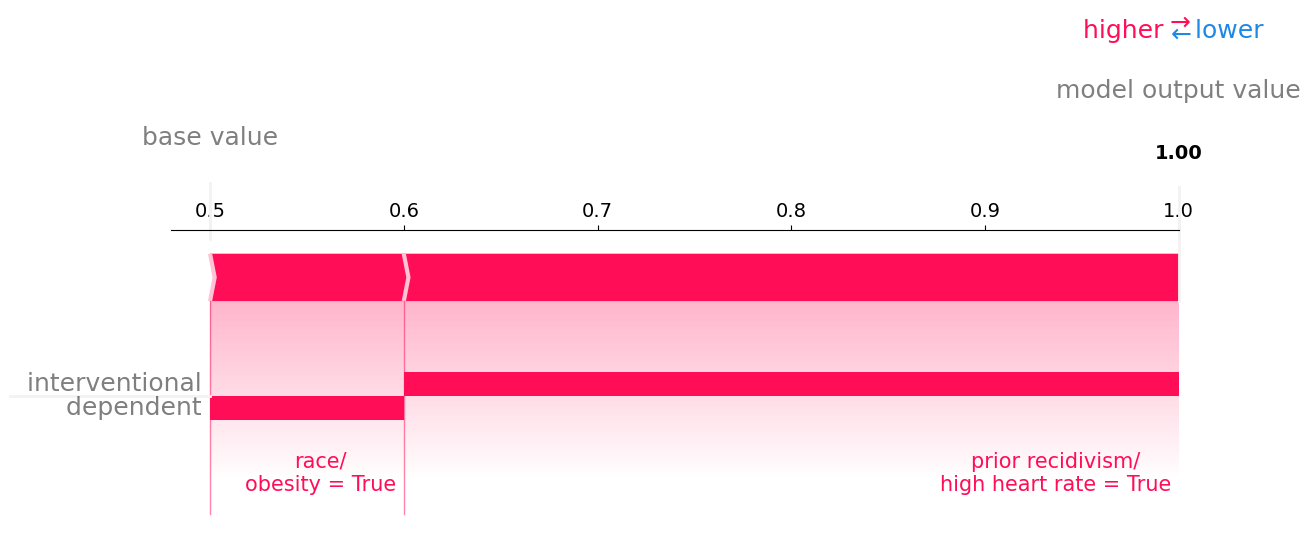}
		\caption{Force plot on sample from the introduction dataset. As can be seen, race or obesity has no direct interventional effect on the model output. They do have an indirect dependent effect via their dependent features, prior recidivism and high heart rate respectively.}
		\label{fig:sim}
	\end{figure}

	Returning to the toy example in the introduction, with $X_1$ and $X_2$ both Bernoulli distributed with $p=0.5$ and $P(X_1=X_2) = 0.7$, we can exactly compute the SHAP parts. For the sample output $f_r(X_1=1,X_2=1) =1$ they are shown as a force plot \cite{lundberg2018explainable} in Figure \ref{fig:sim}. We see that the convict's race has no interventional effect. Whether the patient is obese does have an output effect via its dependent feature `high heart rate' which is clearly shown in the explanation.
	%\begin{center}
	%	\begin{tabular}{ c || c | c}
		%		& $X_1=1$ & $X_2=1$ \\
		%		\hline
		%		\textbf{Interventional SHAP} & 0.5 & 0 \\
		%		\hline
		%		\textbf{Conditional SHAP} & 0.4 & 0.1 \\
		%		\hline
		%		Independent part & 0.4 & 0 \\
		%		\hline
		%		Dependent part & 0 & 0.1 
		%	\end{tabular}
	%\end{center}
	
	We continue this section by first comparing the interventional and dependent SHAP part to the state-of-the-art, whereafter the combined explanation is examined.
		
	\subsection{Interventional SHAP part}
	To underline the model correspondence of the interventional SHAP part, we conduct a first experiment. \cite{chen2020true} trains a logistic regression model on a large dataset of loans and examines which explanation, interventional or conditional SHAP values, helped a loan applicant in decreasing their risk of default. In case of the Boston Housing dataset \cite{harrison1978hedonic}, we can consider a mayor who likes to increase the median house price in his town. To this extent, the town mayor can set the features with the most negative SHAP values to the mean feature value and check the change in median house price. In this experiment, the explanation is better if the median house price increases faster. Note that it is implicitly assumed that the features can be changed (unlike race for example). This is also assumed by \cite{chen2020true}.
	
\begin{figure}
		\centering
		\subfloat[]
		{
			\includegraphics[width=\linewidth]{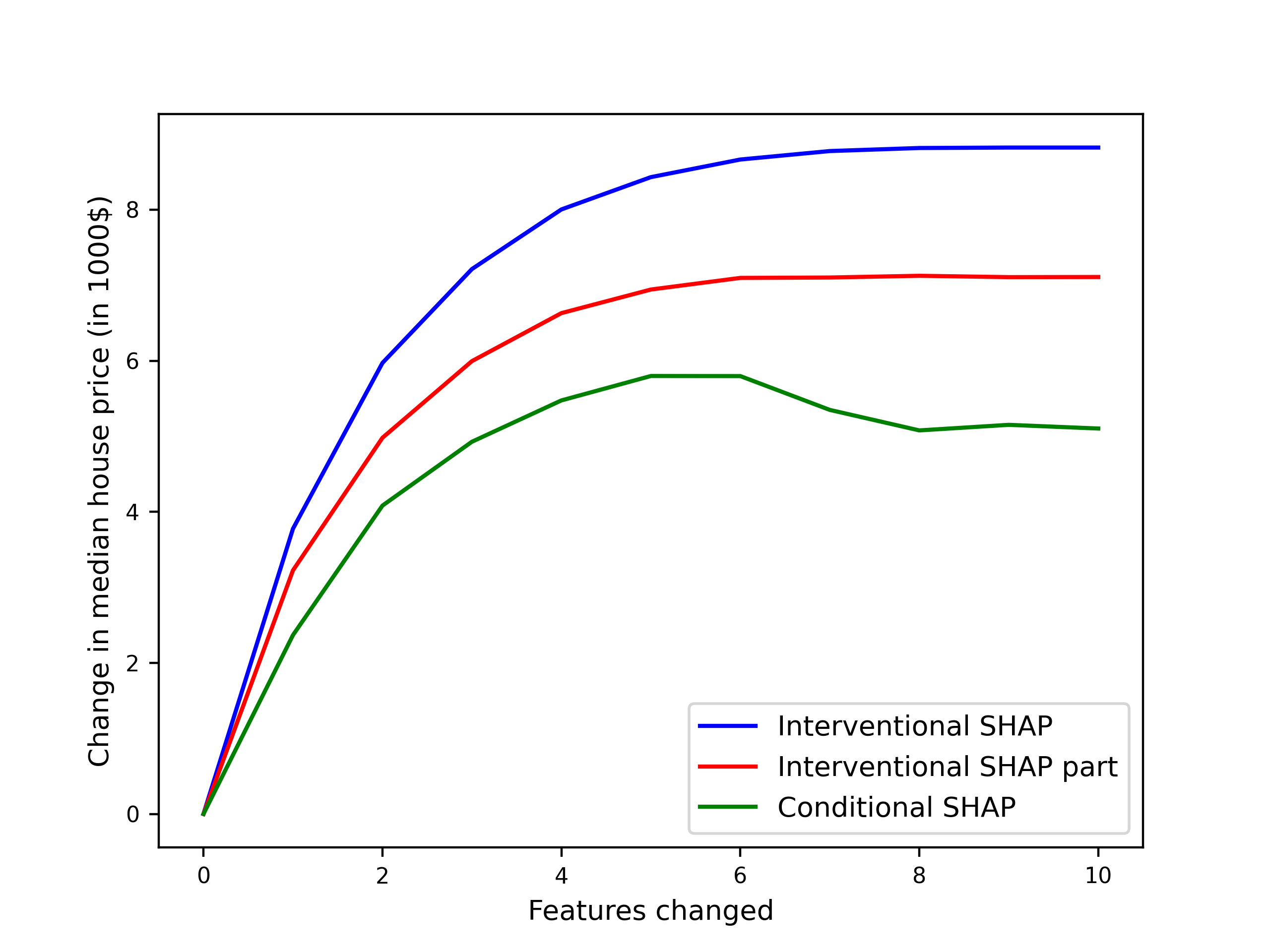}
			\label{fig:linear_diff}} \\
		\subfloat[]
		{
			\includegraphics[width=\linewidth]{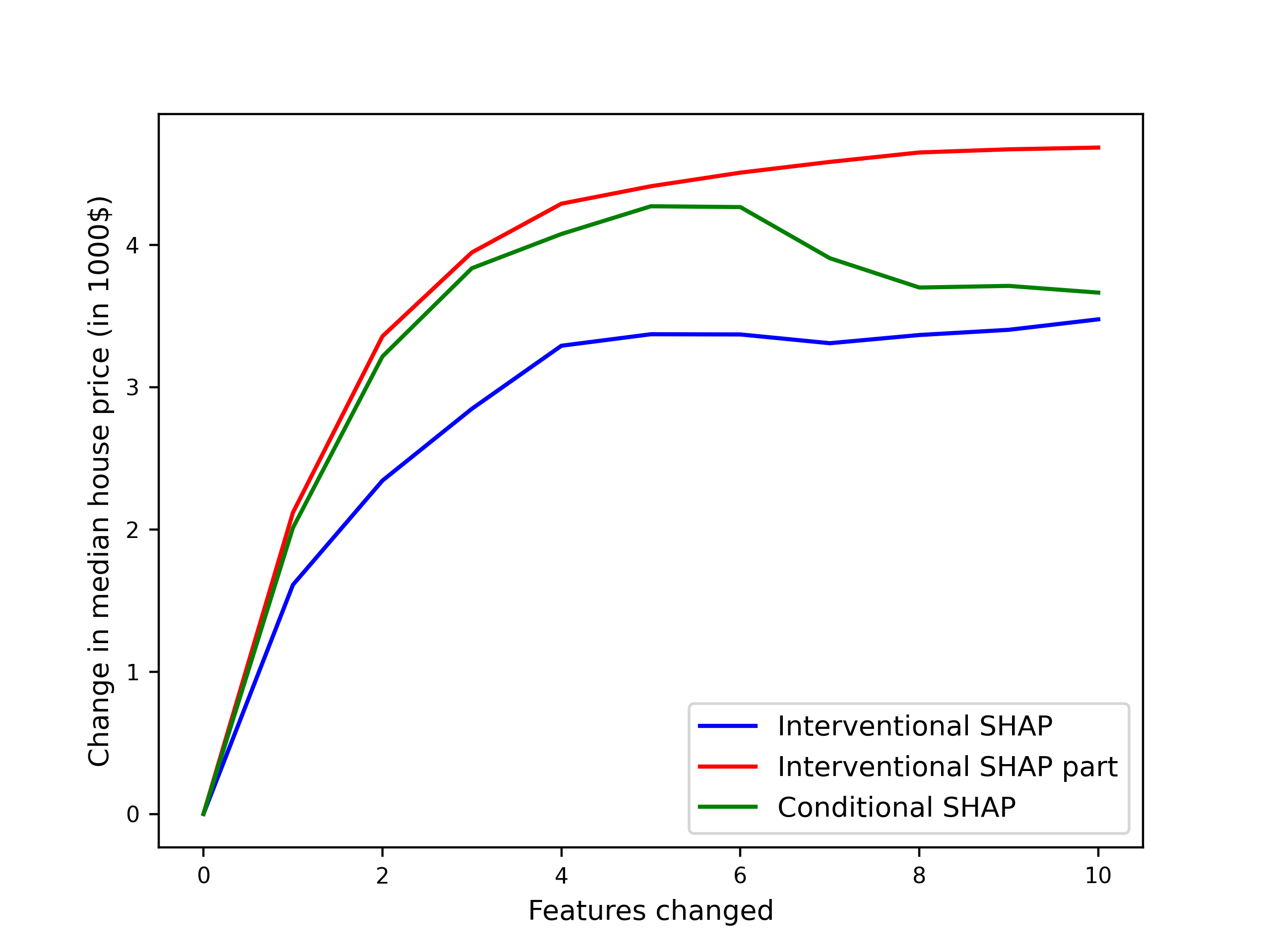}
			\label{fig:linear_diff_c}}
		\caption{Change in house price when imputing selected features with respectively the regular mean (a) and the mean conditioned on the unchanged features (b). A linear model is used to predict the house price. The values are computed for 200 random towns and averaged.}
		\label{fig:linear_diff_all}
	\end{figure}

	In Figure \ref{fig:linear_diff} we show the change in house price when the features are selected by consulting, respectively, the interventional SHAP value, the interventional SHAP parts, the conditional SHAP values. Motivated by the theoretical observations in Section \ref{validation}, a linear model (ordinary least squares) is used. Note that the effective model of \cite{chen2020true} is also linear since the SHAP values are computed on the log odds. The dataset is assumed to be multivariate Gaussian.

	It is also interesting to consider imputing with the mean \textit{conditioned} on the (still) unchanged features. This can be considered a more reasonable way to increase the median house price: how much we can realistically change a feature (e.g. decrease the crime rate) depends on many other features (e.g. the quality of the education in the town measured by the pupil-teacher ratio $\mathit{PTRATIO}$).
	The results are shown in Figure \ref{fig:linear_diff_c}.

	The experiments are also repeated for a non-linear model (a random forest) and are shown in the Appendix.
	
	\subsection{Dependent SHAP part}\label{dependent_experiment}
	Regarding attributions via dependent features we can compare our approach with Shapley residuals \cite{kumar2021shapley}. Consider the following two-dimensional additive model with feature interaction and data distribution:
	\begin{equation}
		\begin{gathered}
		f([X_1, X_2]) = X_1 + X_2 + a_{12} X_1 X_2,\\
		[X_1, X_2] \sim \mathcal{N} \left([0, 0], \left[\begin{array}{ll}
			1 & \alpha \\
			\alpha & 1\end{array}\right] \right).
		\end{gathered}
		\label{eq:additive_model}
	\end{equation}

\begin{figure}
		\centering
		\includegraphics[width=\linewidth]{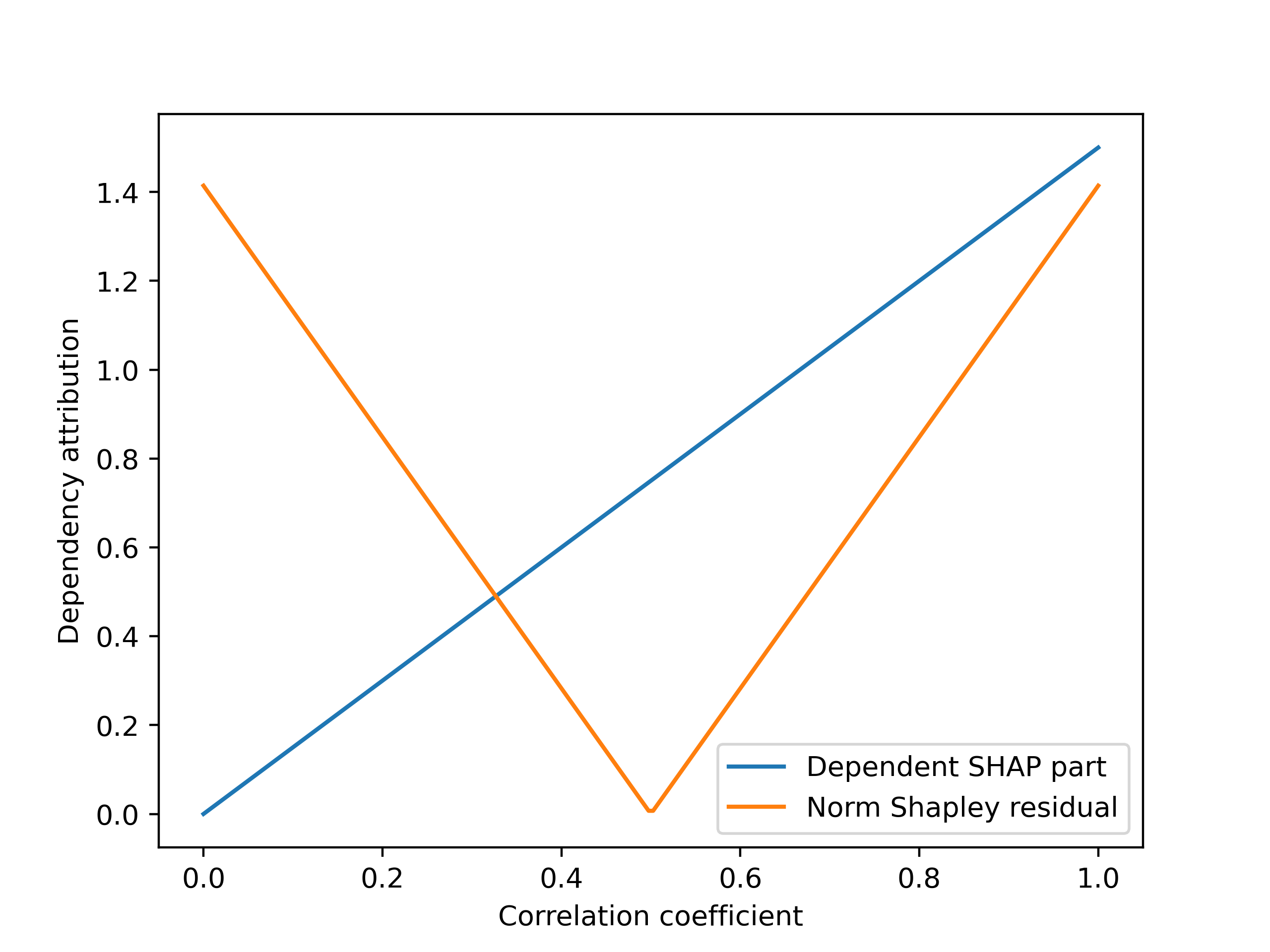}
		
		\caption{The dependency attribution of our method and the state-of-the-art \cite{kumar2021shapley} as a function of the correlation coefficient between two Gaussian variables in an additive model with feature interaction (Equation (\ref{eq:additive_model}) with $a_{12} = 2$).}
		\label{fig:correlation}
	\end{figure}

	For a sample $[X_1=1,X_2=1]$ the conditional SHAP value and dependent SHAP part for $i \in \{1,2\}$ are : 
	\begin{equation}
		\begin{gathered}
			\phi_{i} = 1 + 0.5 a_{12}(1- \alpha), \\
			\phi_{i, dep} = 0.5(1+a_{12}) \alpha.\\
		\end{gathered}
	\end{equation}
	The Shapley residual vector \cite{kumar2021shapley} and its norm are:
	 \begin{equation}\label{eq:residual_example}
	 	\begin{gathered}
	 		r_i = \left[\begin{array}{l}
	 			0.5  a_{12} - (1+0.5a_{12})\alpha \\
	 			- 0.5 a_{12} + (1+0.5a_{12})\alpha
	 		\end{array}\right] \\
 			||r_i|| = \sqrt{2}|0.5  a_{12} - (1+0.5a_{12})\alpha|.
	 	\end{gathered}
	 \end{equation}
 	The relationship of both dependent attributions with respect to the correlation $\alpha$ is shown in Figure \ref{fig:correlation}.

	\subsection{Combined explanation}
\begin{figure}
		\centering
		\includegraphics[width=\linewidth]{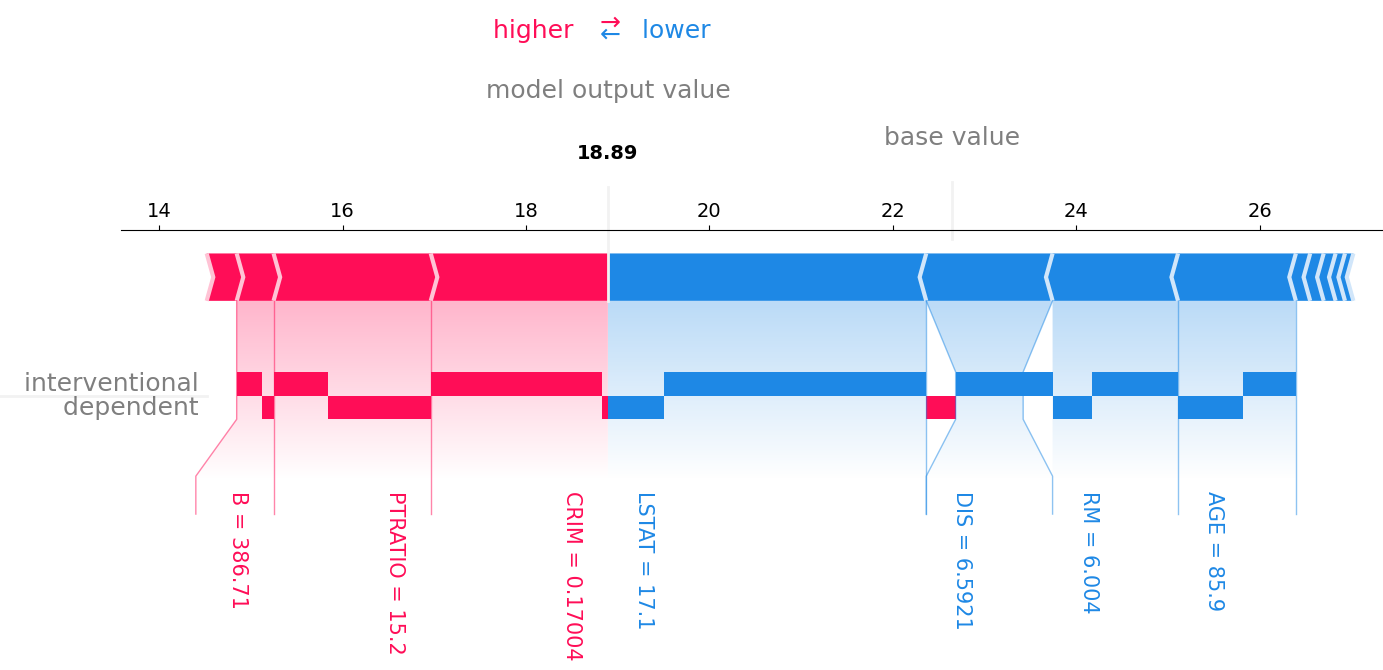}
		\caption{Force plot on sample from Boston Housing dataset}
		\label{fig:boston}
	\end{figure}

	We apply our method again to the Boston Housing dataset \cite{harrison1978hedonic}. This time the model used is a high accuracy random forest and the dataset is assumed to be (multivariate) Gaussian distributed. Figure \ref{fig:boston} depicts one complete explanation, showing a series of interesting local interventional \textit{and} dependent effects. For example, the crime rate per capita ($\mathit{CRIM}$) being 0.17004 has a big positive effect on the median house price in this town, and the effect is largely interventional. This means that the $\mathit{CRIM}$ of this town has a strong effect on the output via its specific weights in the model. The pupil-teacher ratio $\mathit{PTRATIO}$ also has positive effect. A significant part of this effect is through its dependent features: it influences the distributions of the dependent features which in term increase the expected model output via their specific weights. We can also see that the model has a small racial bias: the interventional part of $B$ is non-zero. Note that $B$ is equal to $1000(b-0.63)^2$ with $b$ the black proportion of population \cite{harrison1978hedonic}.

\begin{figure}
		\centering
		\subfloat[]{
			\includegraphics[width=\linewidth]{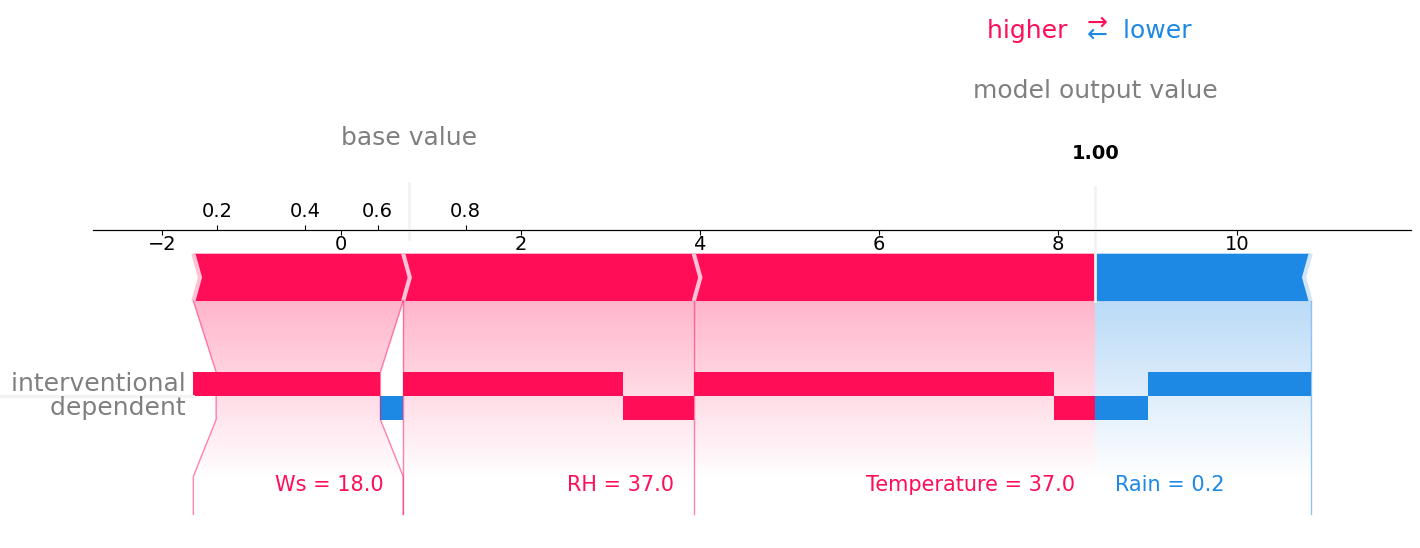}
			\label{fig:fire_correct2}} \\
		\subfloat[]{
			\includegraphics[width=\linewidth]{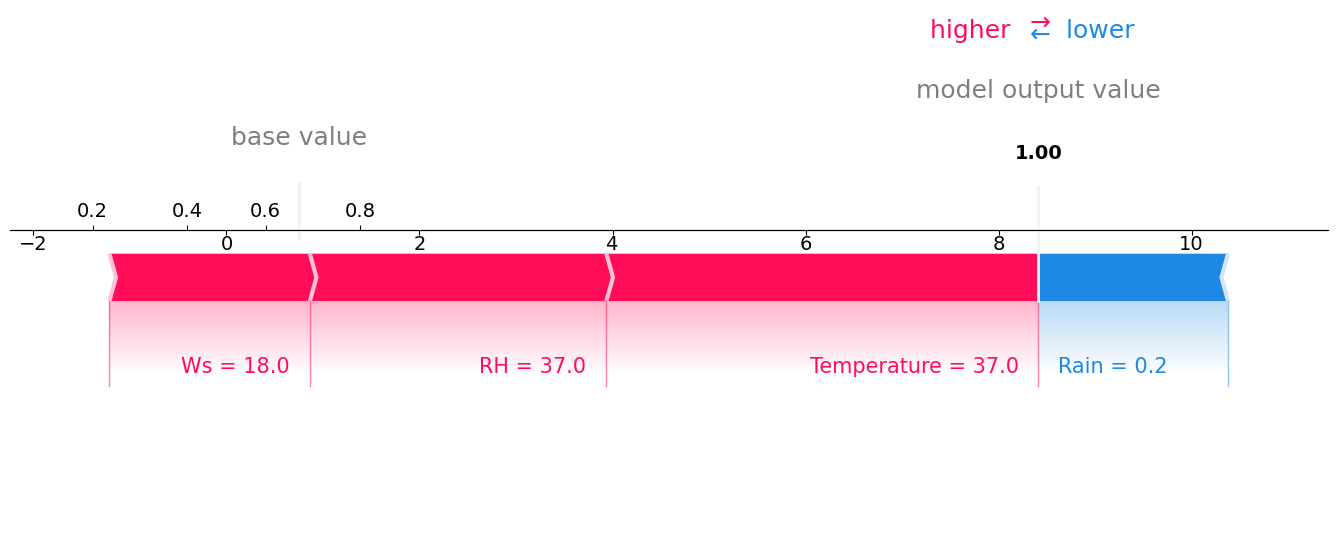}
			\label{fig:fire_correct_int2}}
		\caption{Force plot on a correctly classified sample of the Algerian Forest Fires dataset with our approach (a) and the approach of \cite{lundberg2018explainable} (b) respectively. The top axis denotes the probability (for interpretability) and the one below denotes the log odds (on which the SHAP values and parts are actually computed).}
	\end{figure}

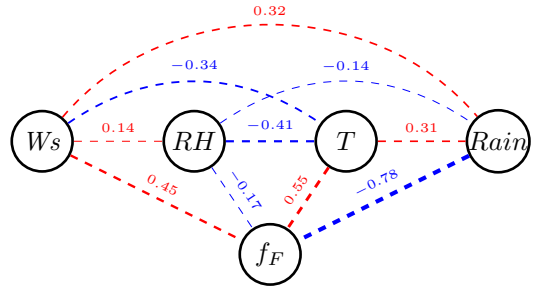
\begin{figure}
		\centering
		\begin{tikzpicture}
			% x node set with absolute coordinates
			\node[state] (RH) at (0,0) {$RH$};
			
			% y node set relative to x.
			% Locations can be:
			% right,left,above,below,
			% above left,below right, etc
			\node[state] (Ws) at (-2,0) {$\mathit{Ws}$};
			\node[state] (Temp)  at (2,0) {$T$};
			\node[state] (Rain) at (4,0) {$Rain$};
			
			\node[state] (f) at (1,-1.5) {$f_F$};
			
			% Association edge

			\path[association] (Ws) edge[red, line width=0.28pt] node[sloped, font=\tiny] {$0.14$} (RH);
			\path[association] (RH) edge[blue, line width=0.82pt]
			node[sloped, font=\tiny] {$-0.41$} (Temp);
			\path[association] (Temp) edge[red, line width=0.62pt]
			node[sloped, font=\tiny] {$0.31$}
			(Rain);
			
			\path[association] (Rain) edge[blue, line width=1.57pt]
			node[sloped, font=\tiny] {$-0.78$} (f);
			\path[association] (Ws) edge[red, line width=0.89pt]
			node[sloped, font=\tiny] {$0.45$} (f);
			\path[association] (RH) edge[blue, line width=0.35pt]
			node[sloped, font=\tiny] {$-0.17$} (f);
			\path[association] (Temp) edge[red, line width=1.10pt]
			node[sloped, font=\tiny] {$0.55$} (f);

			\path[association] (Ws) edge[bend left=35, blue, line width=0.69pt]
			node[sloped, font=\tiny] {$-0.34$} (Temp);
			\path[association] (RH) edge[bend left=35,blue, line width=0.28pt]
			node[sloped, font=\tiny] {$-0.14$} (Rain);
			\path[association] (Ws) edge[bend left=50,red, line width=0.64pt]
			node[sloped, font=\tiny] {$0.32$} (Rain);

		\end{tikzpicture}
		\caption{Partial correlation graph between features and output of the model $f_F$. Spearman's partial correlations are shown on the edges. Red and blue edges respectively show positive and negative correlation. The line width of the edge varies linear with the absolute value of the correlation.}
		\label{fig:partial_corr}
	\end{figure}

	Finally to show that our method generalizes to different datasets and their assumed distribution we apply the method to the Algerian Forest Fires dataset \cite{abid2019predicting} and use a Gaussian copula to model the distribution \cite{aas2019explaining}. We use a random forest to predict the probability of a forest fire based on temperature ($\mathit{T}$), relative humidity ($\mathit{RH}$), wind speed ($\mathit{Ws}$) and $\mathit{Rain}$. A complete explanation can be found in Figure \ref{fig:fire_correct2} with a corresponding classic (interventional) SHAP explanation \cite{lundberg2018explainable} in Figure \ref{fig:fire_correct_int2}. Additionally, in the interest of studying the interventional and dependent connections between features and output, a partial correlation graph was constructed. This allows us to measure the degree of association between two variables, after removing the effect of all other controlling variables. Figure \ref{fig:partial_corr} shows the partial correlation graph for the Algerian Forest Fires dataset, where $f_F$ is the log odds output of the trained model. This graph can be compared with the (normal) correlation between the features and their respective interventional and dependent SHAP parts in Table \ref{shap_corr}. All reported values are Spearman's rank correlations.

	\begin{table}
		\caption{Correlations between feature values and their SHAP values} \label{shap_corr}
		\begin{center}
			\begin{tabular}{r|rrrr}
				\diagbox{$\phi$}{$i$}& $\mathit{Ws}$ & $RH$ & $T$& $Rain$ \\
				\hline \\
				$\phi_{i,int}(f_F)$ & $0.86$ & $-0.69$ & $0.79$ & $-0.88$ \\
				$\phi_{i,dep}(f_F)$ & $0.41$ & $-0.29$ & $-0.52$ & $0.75$ \\
			\end{tabular}
		\end{center}
	\end{table}
	
	\section{Discussion}
	The explanation for the toy example (Figure \ref{fig:sim}) is sufficient in both settings. In the court setting, the convict can clearly see the absence of racial bias. In the hospital setting, it is clear that the patient's obesity has an effect on the risk of a heart attack. 
	
	The other results should validate the interventional and dependent SHAP part and show their equivalence or superiority to the state-of-the-art. Secondly the combined explanation should explain both the model and the result, and should enhance interpretability for different settings and data distributions.
	
	\subsection{Interventional SHAP part}
	From Figure \ref{fig:linear_diff} we can see that on average selecting and mean imputing the features with the biggest negative interventional SHAP part is a better approach to increase the house price than using the conditional SHAP values. Using interventional SHAP values (not parts) still achieves bigger increases in house price, confirming the results of \cite{chen2020true}. Both interventional approaches clearly outperform the conditional SHAP approach, confirming their model correspondence. The Appendix has additional experiments that show the significance of the results. 
	
	Figure \ref{fig:linear_diff} also shows that interventional SHAP values are more true to the model than interventional SHAP parts. This is as expected since for a linear model the exact interventional SHAP value is equal to the change in model output when imputing with the marginal mean (see \eqref{eq:linear_int}). If instead of the marginal, the conditional mean is used, interventional SHAP part comes out on top (Figure \ref{fig:linear_diff_c}). Again, because of \eqref{eq:linear_int_part}, this is an expected result. Note that these observations are independent of the fitting procedure for the linear model: as long as the resulting model is linear, \eqref{eq:linear_int} and \eqref{eq:linear_int_part} will hold and the experiments should lead to the same conclusions. These results show that the interventional SHAP part is indeed a direct model (or interventional) effect and is more representative when the intervention is done with the arguably more intuitive conditional mean. Additionally it adheres to the dummy property (Proposition \ref{prop_dummy}). Thus we can claim that our decomposition provides equivalent interventional insight as interventional SHAP \cite{janzing2020feature}.

	For a non-linear model, \eqref{eq:linear_int} and \eqref{eq:linear_int_part} do not hold. Although the differences between the approaches are less significant in case of a random forest (see the Appendix), they are still in line with the results for a linear model.
	
	\subsection{Dependent SHAP part}
	Figure \ref{fig:correlation}  shows that the dependent SHAP part increases linear with the correlation and is zero when the features are independent, as desired. The norm of the Shapley residual follows a less intuitive trajectory: it is not zero for independent features and does not monotonically increase with the correlation coefficient. Even more, it is zero at a certain non-zero correlation. Shapley residuals for conditional SHAP values also account for feature interactions in the model not only for interactions in the data, a observation not made in \cite{kumar2021shapley}. To effectively explain the model \textit{and} the result, the dependent attribution should not attribute importance solely to feature interactions in the model (which Shapley residuals do for $\alpha=0$). Therefore it is clear that our method is preferred. Furthermore, Shapley residuals can not be interpreted as additive parts of Shapley values: as claimed by Proposition \ref{prop_residual} and confirmed by summing the components of $r_i$ in Equation \eqref{eq:residual_example} this part would always be zero. This makes residuals unsuitable for intuitive representation through force plots, unlike our method.

	\subsection{Combined explanation}
	The explanation for the sample of the Algerian Forest Fire dataset (\ref{fig:fire_correct2}) allows to validate the model output: a lay person would agree with the large interventional effect of the temperature being $37$ degrees Celsius. Furthermore, the combined explanation of interventional and dependent parts allow to motivate certain actions to change the expected result (fire or no fire). We can read that dropping water on this forest should directly decrease the probability of a fire (high interventional effects of relative humidity $\mathit{RH}$ and $\mathit{Rain}$). A smaller (and possible delayed) but significant dependent effect is also expected. This information can not be distilled with the approach of \cite{lundberg2018explainable} (Figure \ref{fig:fire_correct_int2}).
	
	The correlations between features and their SHAP values in Table \ref{shap_corr} are supported by the partial correlations between features and output shown in Figure \ref{fig:partial_corr}. Table \ref{shap_corr} and Figure \ref{fig:partial_corr} both provide a limited view into the model-data structure: they are expected to show similarities but are not comparable in an absolute sense. The partial correlations between each input and the output $f_F$ resemble the correlations between each input and its interventional SHAP part. Furthermore, the big positive correlation of $Rain$ with its dependent SHAP part (in contrast with its big negative correlation with its interventional SHAP part) can be attributed to positive partial correlations with other features ($T$, $\mathit{Ws}$) and their high positive partial correlation with the output $f_F$. These observations validate the decomposition and show how it can give more insight into the model-data structure than other SHAP implementations. Note that Shapley flow \cite{wang2021shapley} would provide even more insight into the model and data dependencies, but their approach is less practical since it require a causal graph and the explanations can not be represented through the popular force plots \cite{lundberg2018explainable}.
	
	In practice, our method can be used on any dataset (and model) regardless of its size and its number of features, and the computational cost can be controlled by the parameters of the method.

	If no reasonable estimation of the data distribution can be made (either data driven or knowledge driven) and/or the interest is clearly in interventional output effects, interventional SHAP is the method of choice. If sufficient causal knowledge is available it might be interesting to look into causal Shapley values \cite{heskes2020causal} or Shapley flow \cite{wang2021shapley}. In all other cases we advise to use our method, since there is only information gain with regards to (standard) conditional SHAP. An overview of the insights our approach provides compared to related approaches is given in Table \ref{full_comparison}.

	\begin{table*}
		\centering
		\caption{Suitability of our decomposition and related Shapley implementations to simultaneously explain different important aspects of the model-data structure. The greyed-out methods require knowledge of the causal graph. In case of causal Shapley values the parentheses denote the lack of implementation. For Shapley residuals, it is clear from our experiment in Section \ref{dependent_experiment} that they are not always suitable for explaining dependencies in the data.}
	\begin{tabular}{m{0.21\linewidth}  >{\centering\arraybackslash} m{0.17\linewidth}  >{\centering\arraybackslash} m{0.17\linewidth}  >{\centering\arraybackslash}m{0.17\linewidth}  >{\centering\arraybackslash} m{0.17\linewidth} } \label{full_comparison}
		 & model & result (model+data) & interactions in model & dependencies in data  \\
		
		decomposition SHAP & \checkmark & \checkmark & &  \checkmark  \\
		\hline
		\textcolor{gray}{Shapley flow} \cite{wang2021shapley} & \checkmark & \checkmark & &  \checkmark  \\
		\hline
		\textcolor{gray}{causal Shapley values} \cite{heskes2020causal} & (\checkmark) & \checkmark & &  (\checkmark)  \\
		\hline
		Shapley residuals \cite{kumar2021shapley} & &  & \checkmark &(\checkmark) \\
		\hline
		conditional SHAP \cite{aas2019explaining} &  & \checkmark & &   \\
		\hline
		interventional SHAP \cite{janzing2020feature} & \checkmark &  & &   \\

	\end{tabular}
	\end{table*}

	\section{Limitations}
	As is often the case with new explanation methods (as those proposed by \cite{aas2019explaining}, \cite{sundararajan2020many} and \cite{janzing2020feature}), validation is mostly done through theory and (our own) intuition. Further work could consist of discussing the results and also the general method formulation with domain experts.
	
	There are also some shortcomings of the method. First of all, the distinction between model and result might only be apparent to the machine learning researcher. The domain expert who is not familiar with machine learning or even statistics might not be interested in this distinction. Secondly, just like conditional SHAP and unlike interventional SHAP, our method requires conditionally sampled feature values. In our real data experiments we used a multivariate Gaussian distribution and a Gaussian copula. Other approaches are proposed by \cite{aas2019explaining}.
	
	\section{Conclusion}
	This paper started by motivating a new point of view on post-hoc explanations: an apparent distinction between model and result explanations. To remove this burden of choice, we derive a novel method, extending conditional SHAP, to combine and unify both explanations. Through theory and experiments we show that our method provides equivalent interventional insight as interventional SHAP while additionally explaining feature dependencies better that the existing state-of-the art. Finally, we contributed a novel Shapley value implementation and accompanying force plots suitable for a wide selection of settings, without requiring causal knowledge.
	
	\section*{Code availability}
	
%	Python code to calculate SHAP parts and corresponding plots on any dataset will be made available after acceptance. The experiments of this paper will also be included.
	Python code to calculate SHAP parts and corresponding plots is available on request.

	\section*{Appendix A. Implementation with Gaussian data distribution} \label{appendix_gausian}
	
	To compute the interventional and dependent SHAP parts, we need access to the conditional data distribution $p\left(X_{\bar{S}} | X_{S} = \boldsymbol{x}_S\right)$ with $S$ a subset of the features. If we assume a multivariate Gaussian data distribution, the conditional distribution is also Gaussian with mean $\mu_{\bar{S}|S}$ and covariance matrix $\Sigma_{\bar{S}|S}$. Write
	
	\begin{equation}
		\Sigma=\left[\begin{array}{ll}
			\Sigma_{S S} & \Sigma_{S \bar{S}} \\
			\Sigma_{\bar{S} S} & \Sigma_{\bar{S} \bar{S}}
		\end{array}\right] \text{and} \
		\mu = \left[\begin{array}{l}
			\mu_S \\
			\mu_{\bar{S}}
		\end{array}\right]\text{,}
	\end{equation}
	
	then
	\begin{equation}
		\mu_{\bar{S}|S} =\mu_{\bar{S}}+\Sigma_{\bar{S} S} \Sigma_{S S}^{-1}\left(\boldsymbol{x}_{S}-\mu_{S}\right)
	\end{equation}
	
	and
	\begin{equation}	
		\Sigma_{\bar{S} | S}=\Sigma_{\bar{S} \bar{S}}-\Sigma_{\bar{S} S} \Sigma_{S S}^{-1} \Sigma_{S \bar{S}}\text{.}
	\end{equation}
	
	Now the algorithm to compute the SHAP parts can be formulated:
	\begin{algorithmic}[1]
		\STATE \textbf{given} dataset $\mathbf{X}_{N\times M}$, sample to explain $\boldsymbol{x}$, model $f$, parameters $K_1$ and $K_2$
		\STATE compute mean $\mu$ and covariance matrix $\Sigma$ from $\mathbf{X}_{N\times M}$
		\STATE compute the conditional SHAP values $\phi_{i}$ by applying Kernel SHAP, estimating every $E\left[f(X) | X_S = \boldsymbol{x}_{S}\right] \approx \frac{1}{K_1} \sum_{k}f\left(\left[\boldsymbol{x}_S, x^k_{\bar{S}}\right]\right)$ with $x^k_{\bar{S}}$ sampled from \newline $\mathcal{N}(\mu_{\bar{S}|S}, \Sigma_{\bar{S}|S})$.
		\FOR{$i = 0$ to $M$}
		\FOR{$k = 0$ to $K_2$}
		\STATE $R \leftarrow \text{Permutation}(\{1,2, \dots M\})$
		\STATE sample $x^0_{\bar{S^R}}$ from $\mathcal{N}(\mu_{\bar{S^R}|S^R}, \Sigma_{\bar{S^R}|S^R})$
		\ENDFOR
		\STATE $\phi_{i, int} \leftarrow \frac{1}{K_2} \sum_{R}f\left(\left[\boldsymbol{x}_{S^R}, \boldsymbol{x}_i, x^0_{\bar{S^R_i}}\right]\right) - f\left(\left[\boldsymbol{x}_{S^R}, x^0_{\bar{S^R}}\right]\right)$
		\STATE $\phi_{i, dep} \leftarrow \phi_{i} - \phi_{i, int}$
		\ENDFOR
	\end{algorithmic}
	A lengthy description of the Kernel SHAP algorithm is available in \cite{aas2019explaining}. Lines 4 to 9 compute the interventional SHAP parts in the same way as the normal SHAP values were computed in \cite{vstrumbelj2014explaining}.
	
	\section*{Appendix B. Results non-linear model}
\begin{figure}[h]
		\centering
		\subfloat[]{
			\includegraphics[width=\linewidth]{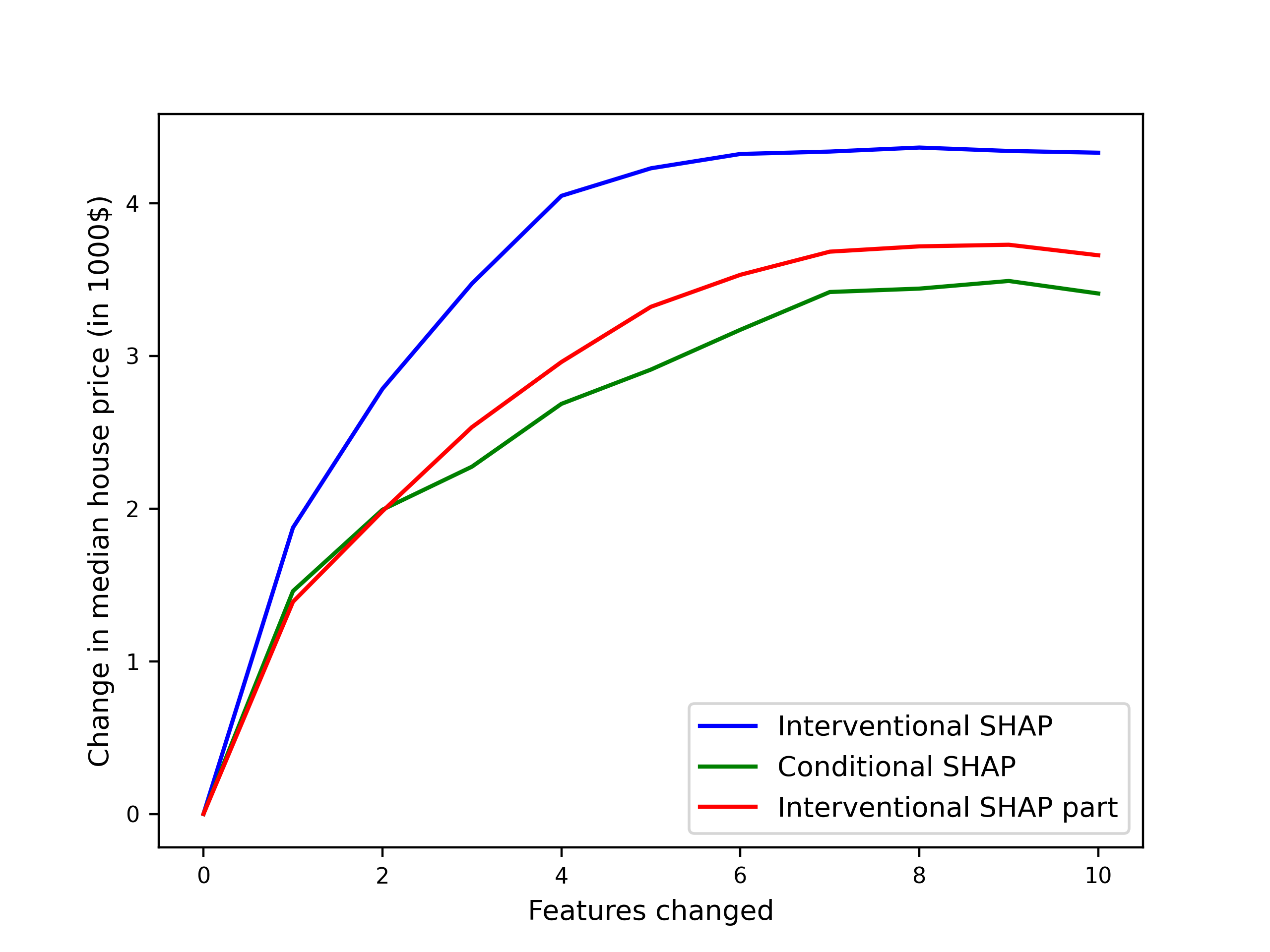}
			\label{fig:forest_diff}} \\
		\subfloat[]{
		\includegraphics[width=\linewidth]{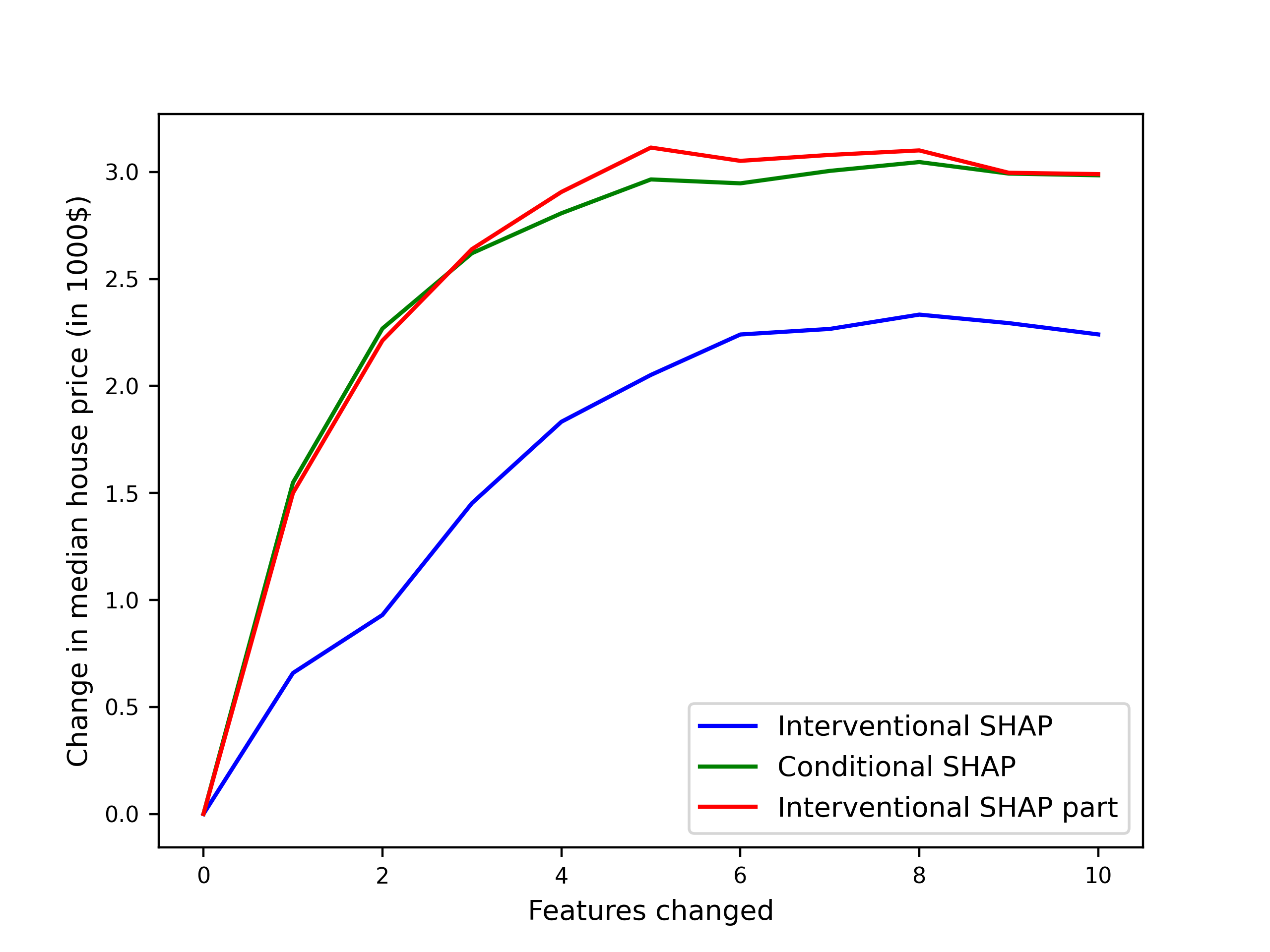}
			\label{fig:forest_diff_c}}
		\caption{Change in house price when imputing selected features with respectively the regular mean (a) and the mean conditioned on the unchanged features (b). A random forest is used to predict the house price. The values are computed for 200 random towns and averaged.}
		\label{fig:forest_diff_all}
	\end{figure}

	Because of Equations \eqref{eq:linear_int} and \eqref{eq:linear_int_part}, interpretation of the results with a linear model is more straightforward. For completeness' sake, we shown the results with a non-linear model here. The experiments of Figure \ref{fig:linear_diff_all}  were repeated with a random forest in Figure \ref{fig:forest_diff_all}. As discussed above, while the differences between the approaches are less significant, they are still in line with the results for a linear
	model.
	
	\section*{Appendix C. Significance of results}
\begin{figure}[h]
		\centering
		\includegraphics[width=\linewidth]{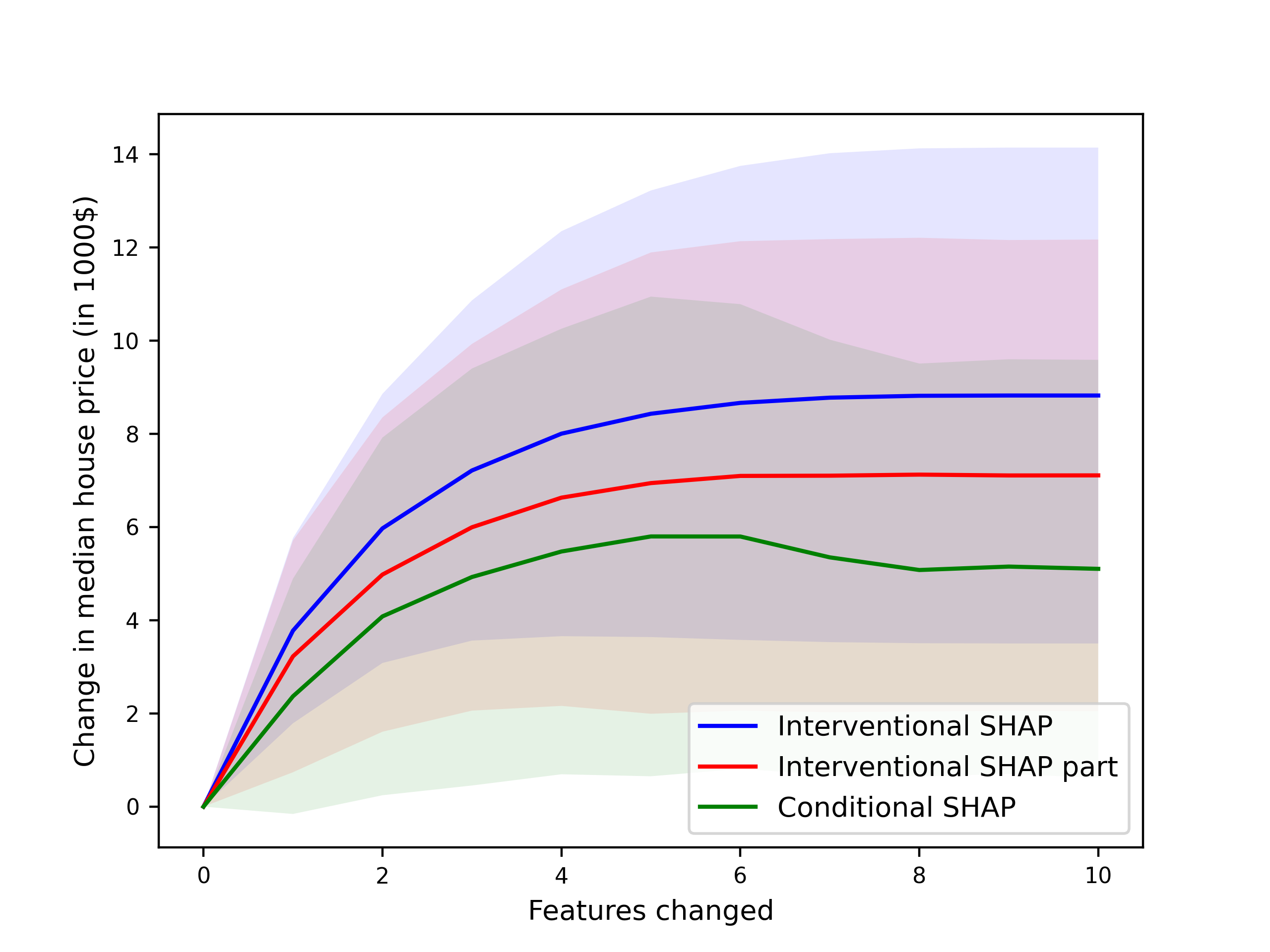}
		\caption{Change in house price when mean imputing selected features. A linear model is used to predict the house price. The values are computed for 200 random towns and averaged. The shaded region represents one standard deviation.}
		\label{fig:linear_diff_std}
	\end{figure}

\begin{figure}[h]
		\centering
		\subfloat[]{
			\includegraphics[width=\linewidth]{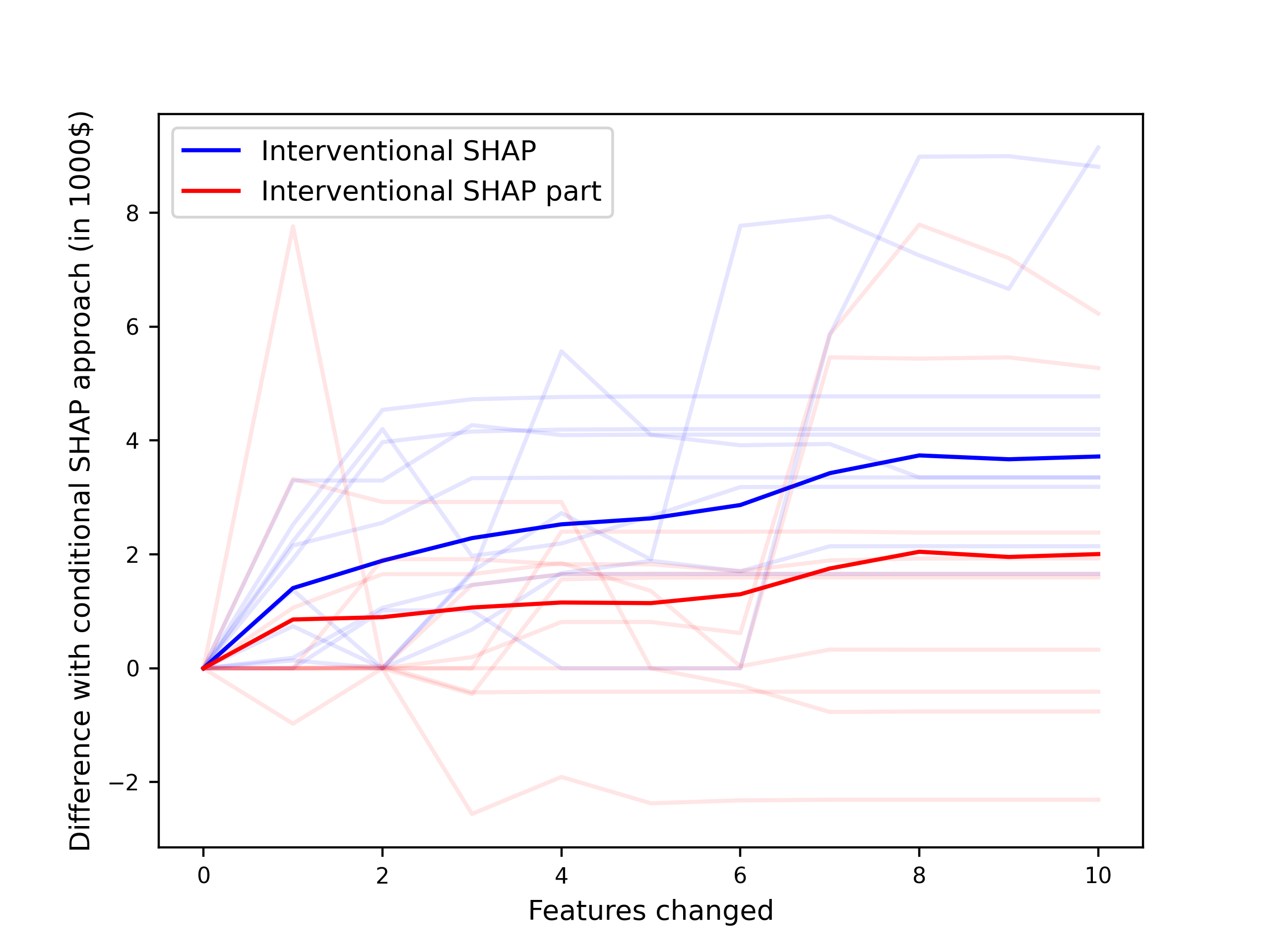}
			\label{fig:linear_norm}} \\
		\subfloat[]{
		\includegraphics[width=\linewidth]{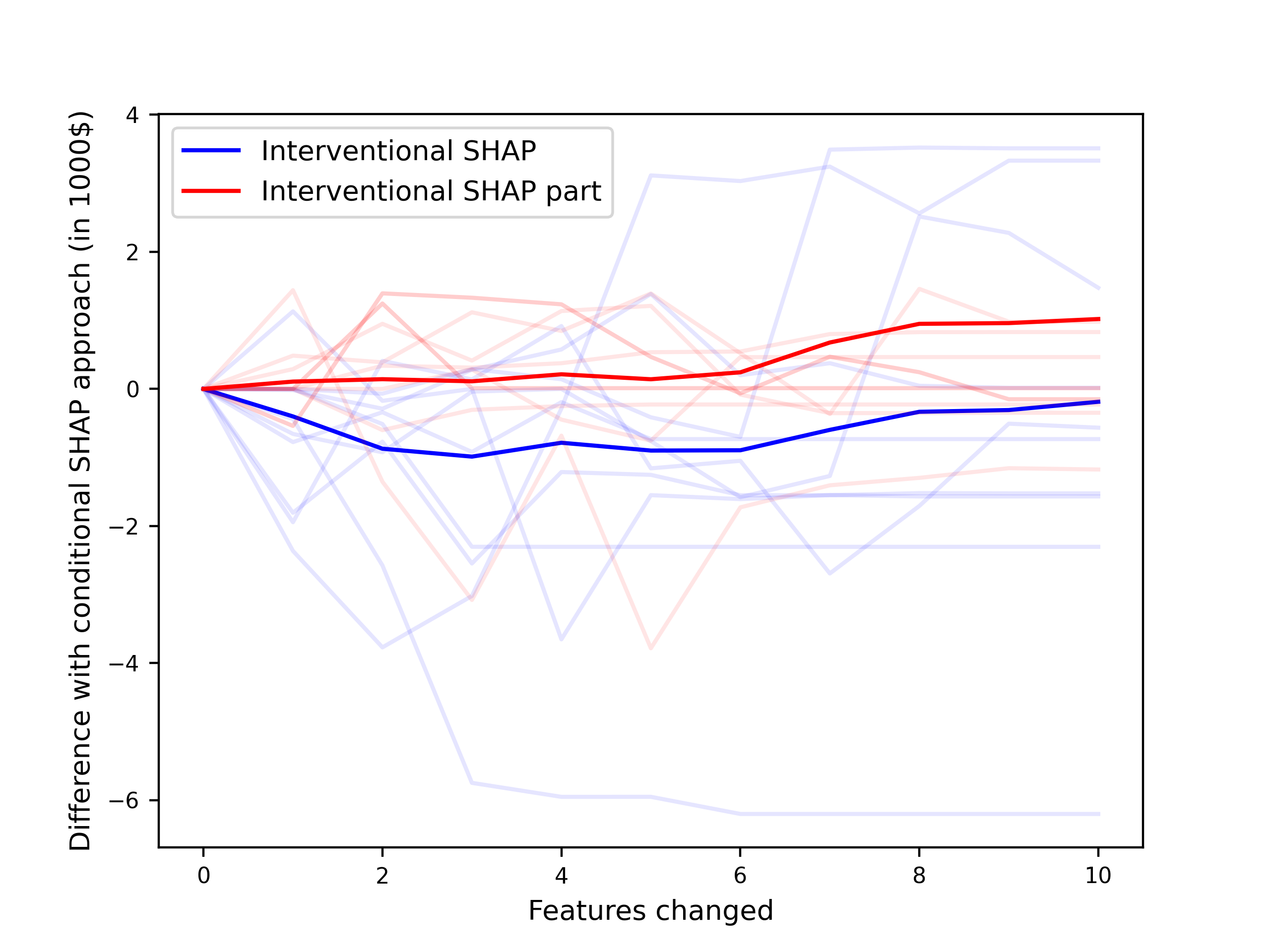}
			\label{fig:linear_norm_c}}
		\caption{The per town difference of both interventional approaches with conditional SHAP when imputing selected features with respectively the regular mean (a) and the mean conditioned on the unchanged features (b). A linear model is used to predict the house price. The mean over 200 towns is a solid line. For ten randomly selected towns, actual differences are plotted (see-through lines).}
		\label{fig:linear_norm_all}
	\end{figure}

\begin{figure}[h!]
		\centering
		\subfloat[]{
			\includegraphics[width=\linewidth]{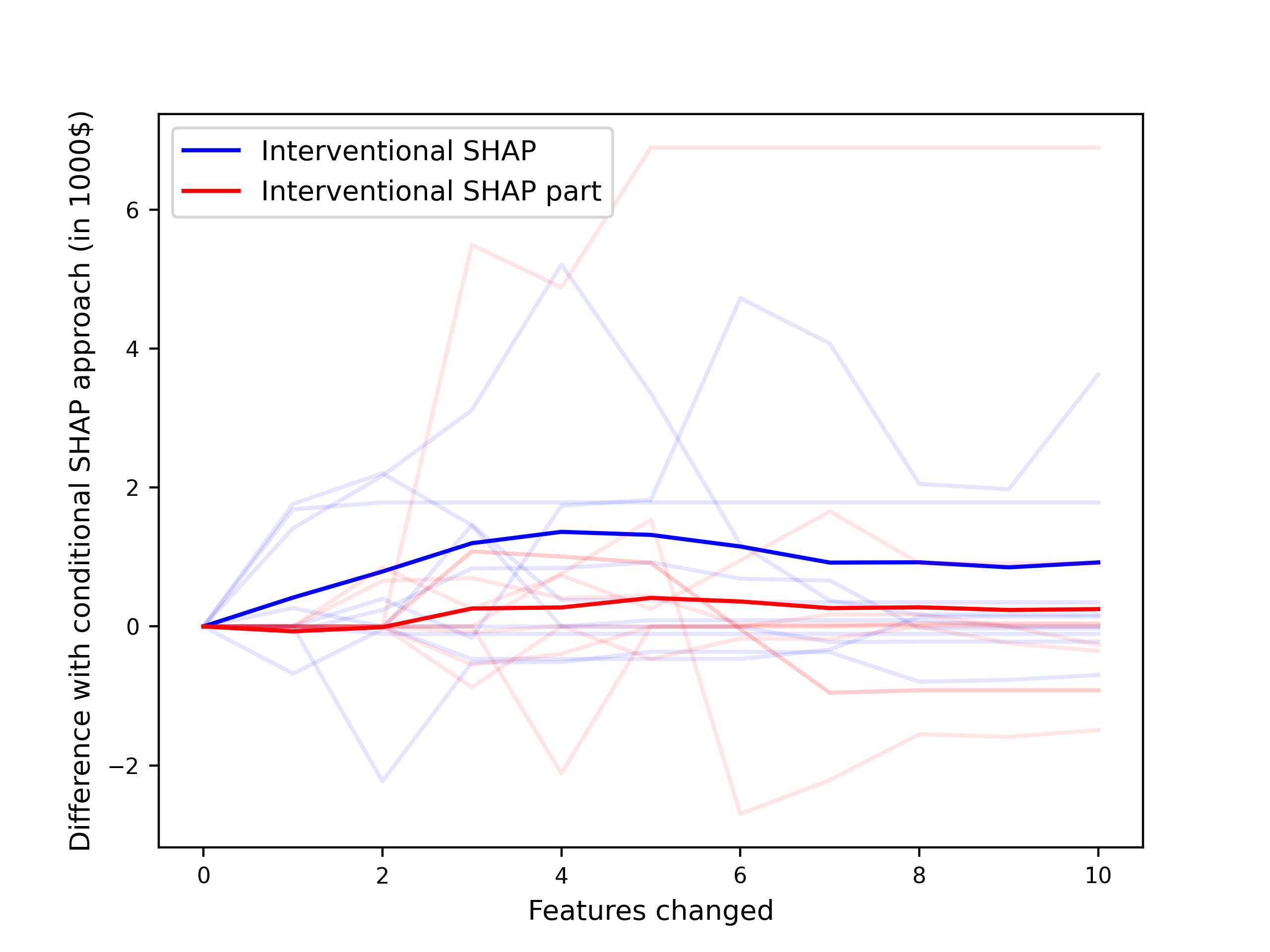}
			\label{fig:forest_norm}} \\
		\subfloat[]
		{\includegraphics[width=\linewidth]{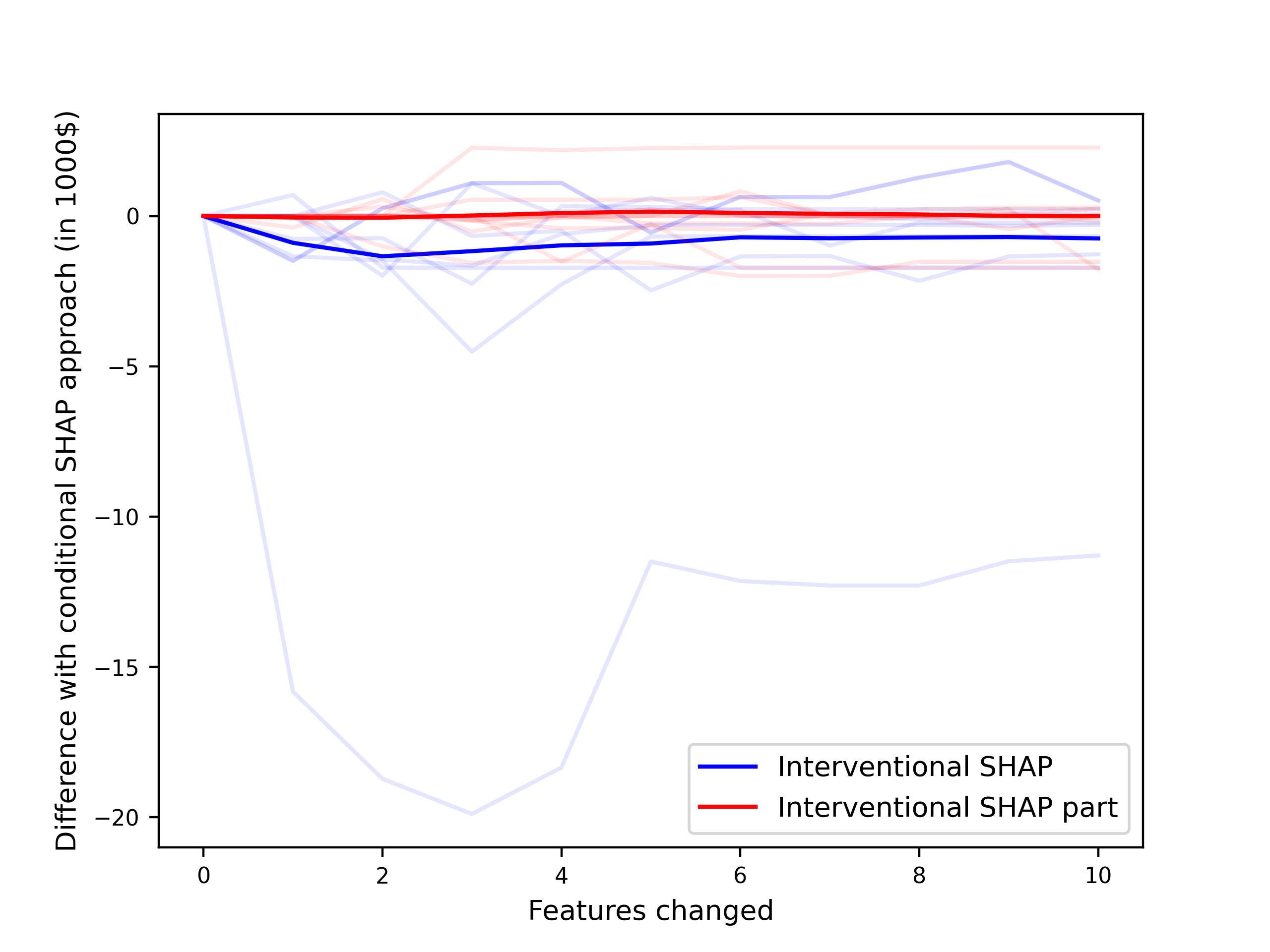}
			\label{fig:forest_norm_c}}
		\caption{The per town difference of both interventional approaches with conditional SHAP when imputing selected features with respectively the regular mean (a) and the mean conditioned on the unchanged features (b). A random forest is used to predict the house price. The mean over 200 towns is a solid line. For ten randomly selected towns, actual differences are plotted (see-through lines).}
		\label{fig:forest_norm_all}
	\end{figure}

	Contrary to regular machine learning practice the averages in Figure \ref{fig:linear_diff_all} and Figure \ref{fig:forest_diff_all} are not accompanied with standard deviations. This was a deliberate choice. Since some towns are very difficult to improve while others can be improved quite easily, the standard deviations are very
	large and make the results seem insignificant (see Figure \ref{fig:linear_diff_std}). A better metric to judge the significance is the
	per town difference of both interventional approaches with conditional SHAP depicted in Figure \ref{fig:linear_norm_all} (for a linear model) and Figure \ref{fig:forest_norm_all} (for a random forest). Here, in addition to the mean we show the actual differences for some towns instead of a standard deviation because these values are far from Gaussian. We can see a significant difference between the approaches.
	
	\bibliography{ref}

\begin{thebibliography}{}

\bibitem[Aas et~al., 2021a]{aas2019explaining}
Aas, K., Jullum, M., and Løland, A. (2021a).
\newblock Explaining individual predictions when features are dependent: More
  accurate approximations to shapley values.
\newblock {\em Artificial Intelligence}, 298:103502.

\bibitem[Aas et~al., 2021b]{aas2021explaining}
Aas, K., Nagler, T., Jullum, M., and L{\o}land, A. (2021b).
\newblock Explaining predictive models using shapley values and non-parametric
  vine copulas.
\newblock {\em Dependence Modeling}, 9(1):62--81.

\bibitem[Abid and Izeboudjen, 2019]{abid2019predicting}
Abid, F. and Izeboudjen, N. (2019).
\newblock Predicting forest fire in algeria using data mining techniques: Case
  study of the decision tree algorithm.
\newblock In {\em International Conference on Advanced Intelligent Systems for
  Sustainable Development}, pages 363--370. Springer.

\bibitem[Chen et~al., 2020]{chen2020true}
Chen, H., Janizek, J.~D., Lundberg, S., and Lee, S.-I. (2020).
\newblock True to the model or true to the data?
\newblock {\em arXiv preprint arXiv:2006.16234}.

\bibitem[Frye et~al., 2020a]{frye2020shapley}
Frye, C., de~Mijolla, D., Begley, T., Cowton, L., Stanley, M., and Feige, I.
  (2020a).
\newblock Shapley explainability on the data manifold.
\newblock {\em arXiv preprint arXiv:2006.01272}.

\bibitem[Frye et~al., 2020b]{frye2020asymmetric}
Frye, C., Rowat, C., and Feige, I. (2020b).
\newblock Asymmetric shapley values: incorporating causal knowledge into
  model-agnostic explainability.
\newblock {\em Advances in Neural Information Processing Systems},
  33:1229--1239.

\bibitem[Harrison~Jr and Rubinfeld, 1978]{harrison1978hedonic}
Harrison~Jr, D. and Rubinfeld, D.~L. (1978).
\newblock Hedonic housing prices and the demand for clean air.
\newblock {\em Journal of environmental economics and management},
  5(1):81--102.

\bibitem[Heskes et~al., 2020]{heskes2020causal}
Heskes, T., Sijben, E., Bucur, I.~G., and Claassen, T. (2020).
\newblock Causal shapley values: Exploiting causal knowledge to explain
  individual predictions of complex models.
\newblock {\em Advances in Neural Information Processing Systems}, 33.

\bibitem[Janzing et~al., 2020]{janzing2020feature}
Janzing, D., Minorics, L., and Bl{\"o}baum, P. (2020).
\newblock Feature relevance quantification in explainable ai: A causal problem.
\newblock In {\em International Conference on Artificial Intelligence and
  Statistics}, pages 2907--2916. PMLR.

\bibitem[Kumar et~al., 2021]{kumar2021shapley}
Kumar, I., Scheidegger, C., Venkatasubramanian, S., and Friedler, S. (2021).
\newblock Shapley residuals: Quantifying the limits of the shapley value for
  explanations.
\newblock {\em Advances in Neural Information Processing Systems},
  34:26598--26608.

\bibitem[Kumar et~al., 2020]{kumar2020problems}
Kumar, I.~E., Venkatasubramanian, S., Scheidegger, C., and Friedler, S. (2020).
\newblock Problems with shapley-value-based explanations as feature importance
  measures.
\newblock In {\em International Conference on Machine Learning}, pages
  5491--5500. PMLR.

\bibitem[Lockwood et~al., 2015]{lockwood2015racial}
Lockwood, S.~K., Nally, J.~M., Ho, T., and Knutson, K. (2015).
\newblock Racial disparities and similarities in post-release recidivism and
  employment among ex-prisoners with a different level of education.
\newblock {\em Journal of Prison Education and Reentry}, 2(1):16--31.

\bibitem[Lundberg et~al., 2018a]{lundberg2018consistent}
Lundberg, S.~M., Erion, G.~G., and Lee, S.-I. (2018a).
\newblock Consistent individualized feature attribution for tree ensembles.
\newblock {\em arXiv preprint arXiv:1802.03888}.

\bibitem[Lundberg and Lee, 2017]{lundberg2017unified}
Lundberg, S.~M. and Lee, S.-I. (2017).
\newblock A unified approach to interpreting model predictions.
\newblock In {\em Advances in Neural Information Processing Systems}, pages
  4765--4774.

\bibitem[Lundberg et~al., 2018b]{lundberg2018explainable}
Lundberg, S.~M., Nair, B., Vavilala, M.~S., Horibe, M., Eisses, M.~J., Adams,
  T., Liston, D.~E., Low, D. K.-W., Newman, S.-F., Kim, J., et~al. (2018b).
\newblock Explainable machine-learning predictions for the prevention of
  hypoxaemia during surgery.
\newblock {\em Nature Biomedical Engineering}, 2(10):749.

\bibitem[Ribeiro et~al., 2016]{ribeiro2016should}
Ribeiro, M.~T., Singh, S., and Guestrin, C. (2016).
\newblock " why should i trust you?" explaining the predictions of any
  classifier.
\newblock In {\em Proceedings of the 22nd ACM SIGKDD international conference
  on knowledge discovery and data mining}, pages 1135--1144.

\bibitem[Rossi et~al., 2015]{rossi2015impact}
Rossi, R.~C., Vanderlei, L. C.~M., Gon{\c{c}}alves, A. C. C.~R., Vanderlei,
  F.~M., Bernardo, A. F.~B., Yamada, K. M.~H., da~Silva, N.~T., and de~Abreu,
  L.~C. (2015).
\newblock Impact of obesity on autonomic modulation, heart rate and blood
  pressure in obese young people.
\newblock {\em Autonomic neuroscience}, 193:138--141.

\bibitem[Shapley, 1953]{shapley1953value}
Shapley, L.~S. (1953).
\newblock A value for n-person games.
\newblock {\em Contributions to the Theory of Games}, 2(28):307--317.

\bibitem[Slack et~al., 2020]{slack2020fooling}
Slack, D., Hilgard, S., Jia, E., Singh, S., and Lakkaraju, H. (2020).
\newblock Fooling lime and shap: Adversarial attacks on post hoc explanation
  methods.
\newblock In {\em Proceedings of the AAAI/ACM Conference on AI, Ethics, and
  Society}, pages 180--186.

\bibitem[{\v{S}}trumbelj and Kononenko, 2014]{vstrumbelj2014explaining}
{\v{S}}trumbelj, E. and Kononenko, I. (2014).
\newblock Explaining prediction models and individual predictions with feature
  contributions.
\newblock {\em Knowledge and information systems}, 41(3):647--665.

\bibitem[Sundararajan et~al., 2020]{sundararajan2020shapley}
Sundararajan, M., Dhamdhere, K., and Agarwal, A. (2020).
\newblock The shapley taylor interaction index.
\newblock In {\em International conference on machine learning}, pages
  9259--9268. PMLR.

\bibitem[Sundararajan and Najmi, 2020]{sundararajan2020many}
Sundararajan, M. and Najmi, A. (2020).
\newblock The many shapley values for model explanation.
\newblock In {\em International Conference on Machine Learning}, pages
  9269--9278. PMLR.

\bibitem[Takeishi, 2019]{takeishi2019shapley}
Takeishi, N. (2019).
\newblock Shapley values of reconstruction errors of pca for explaining anomaly
  detection.
\newblock In {\em 2019 International Conference on Data Mining Workshops
  (ICDMW)}, pages 793--798. IEEE.

\bibitem[Wang et~al., 2021]{wang2021shapley}
Wang, J., Wiens, J., and Lundberg, S. (2021).
\newblock Shapley flow: A graph-based approach to interpreting model
  predictions.
\newblock In {\em International Conference on Artificial Intelligence and
  Statistics}, pages 721--729. PMLR.

\bibitem[Yeh et~al., 2022]{yeh2022threading}
Yeh, C.-K., Lee, K.-Y., Liu, F., and Ravikumar, P. (2022).
\newblock Threading the needle of on and off-manifold value functions for
  shapley explanations.
\newblock In {\em International Conference on Artificial Intelligence and
  Statistics}, pages 1485--1502. PMLR.

\end{thebibliography}
	
\end{document}